\documentclass{article}

\usepackage{microtype}
\usepackage{graphicx}
\usepackage{subcaption}
\usepackage{booktabs} 

\usepackage{hyperref}

\usepackage[accepted]{icml2026}

\usepackage{amsmath}
\usepackage{amssymb}
\usepackage{mathtools}
\usepackage{amsthm}

\usepackage[capitalize,noabbrev]{cleveref}

\theoremstyle{plain}

\theoremstyle{definition}

\theoremstyle{remark}

\usepackage[textsize=tiny]{todonotes}

\usepackage{pdfrender}
\newcommand*{\boldcheckmark}{%
  \textpdfrender{
    TextRenderingMode=FillStroke,
    LineWidth=1pt, 
  }{\checkmark}%
}

\icmltitlerunning{Infinite-Precision Autoregressive Modeling for Vector Graphics and Layouts}

\begin{document}

\twocolumn[
  \icmltitle{Infinite-Precision Autoregressive Modeling for Vector Graphics and Layouts}



  \icmlsetsymbol{equal}{*}

  \begin{icmlauthorlist}
    \icmlauthor{Yeonsang Shin}{equal,snu1} \quad
    \icmlauthor{Insoo Kim}{equal,snu1} \quad
    \icmlauthor{Bongkeun Kim}{samsung} \quad
    \icmlauthor{Keonwoo Bae}{samsung} \quad
    \icmlauthor{Bohyung Han}{snu1,snu2}
  \end{icmlauthorlist}

  \icmlaffiliation{snu1}{ECE, Seoul National University, Korea}
  \icmlaffiliation{samsung}{Samsung Electronics Co., Korea}
  \icmlaffiliation{snu2}{IPAI, Seoul National University, Korea}

  \icmlcorrespondingauthor{Bohyung Han}{bhhan@snu.ac.kr}

  \icmlkeywords{joint discrete-continuous generation, autoregression, high-precision, layout generation, svg}

  \vskip 0.3in
]



\printAffiliationsAndNotice{\icmlEqualContribution}


\begin{abstract}

While Transformer-based autoregressive models excel in data generation, their token discretization strategy inherently limits their precision in continuous domains.
We analyze the scalability limitations of existing discretization-based approaches for generating hybrid discrete-continuous sequences, particularly in high-precision domains such as logos, layouts, and semiconductor circuit designs, where precision loss potentially leads to visual artifacts, aesthetic degradation, and even functional failure.
To address the challenge, we propose a novel unified framework that jointly models discrete and continuous values for variable-length sequences.  
Our approach employs a hybrid approach that combines categorical prediction for discrete values with diffusion-based modeling for continuous values, incorporating two key technical components: an end-of-sequence~(EOS) logit adjustment mechanism that uses an MLP to dynamically adjust EOS token logits based on sequence context, and a length regularization term integrated into the loss function.
Additionally, we present ContLayNet, a large-scale benchmark comprising 334K high-precision semiconductor layout samples with specialized evaluation metrics that capture functional correctness, where precision errors significantly impact performance.
%
Experiments on multiple domains show that our approach achieves higher-fidelity hybrid vector representations than discretization-based and fixed-schema baselines, while effectively scaling to high-precision generation.

\end{abstract}

\section{Introduction}
\label{sec:intro}


Autoregressive models~\citep{vaswani2017attention, brown2020languagemodelsfewshotlearners, ramesh2021zeroshottexttoimagegeneration, radford2022robustspeechrecognitionlargescale} have shown remarkable success in generating sequences across various domains, typically relying on discrete values such as text tokens and quantized image representations.
However, many real-world applications inherently require hybrid representations that combine discrete and continuous values in variable-length sequences.
Layout designs integrate discrete component types with precise continuous positions, Scalable Vector Graphics~(SVG) pair discrete drawing commands with continuous coordinates, and semiconductor routings associate discrete layer types with continuous geometries.
%
Each of these systems naturally varies in sequence length depending on content complexity.

\begin{figure}[t]
\centering 
\includegraphics[width=\linewidth]{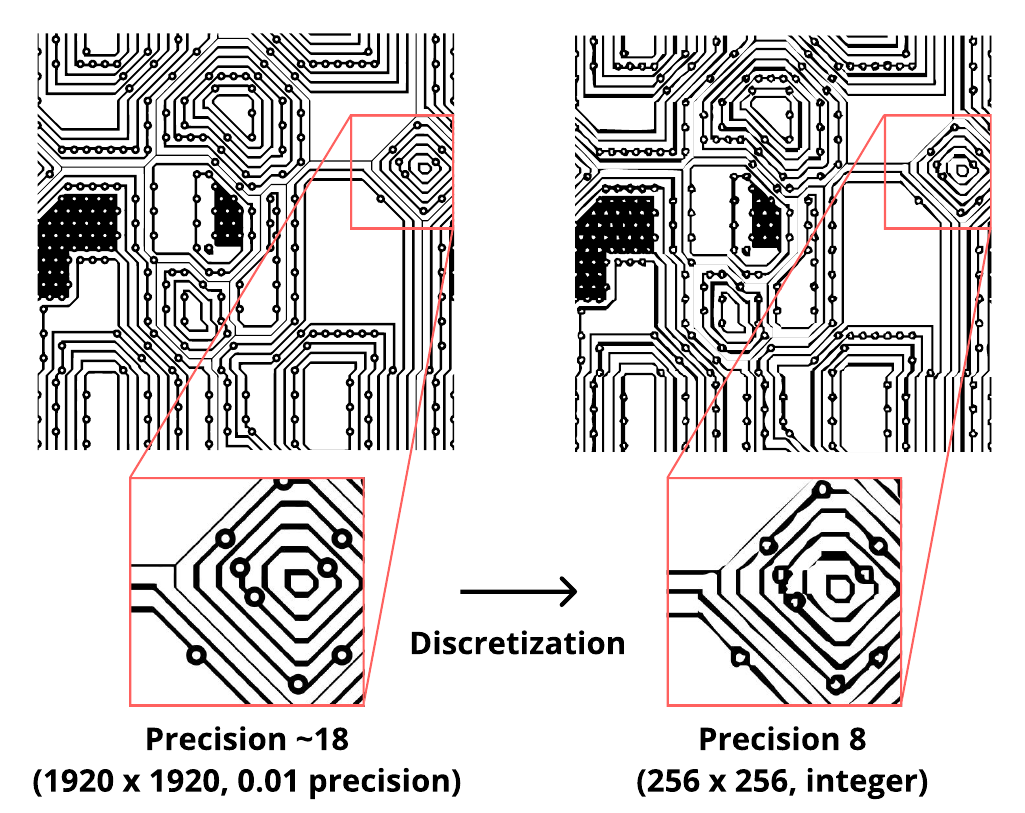}
\caption{\small
\textbf{Impact of discretization.} Discretization fundamentally compromises the scalability of precision. For precision analysis, refer to~\cref{subsec:precision}.
}
\vspace{-3mm}
\label{fig:intro_circuit}
\end{figure}

\begin{table*}[t]
\centering
\setlength{\tabcolsep}{16pt}
{
\caption{\footnotesize
\textbf{Comparison with existing works.}
IPAM is the first autoregressive framework to jointly model hybrid discrete-continuous representations while supporting variable-length sequence generation.
}\label{tab:scope_comparisons}
{
\vspace{-1mm}
\scalebox{0.85}{
\begin{tabular}{l|c|c|c|l}
\toprule
\multicolumn{1}{c|}{} & \multicolumn{1}{c|}{Hybrid?} & \multicolumn{1}{c|}{Autoregressive?} & Variable-Length?       & Domain                  \\ \midrule
LT~\citep{gupta2021layouttransformerlayoutgenerationcompletion}                     &                                                 & \checkmark                                   & \checkmark     & Layout         \\ 
IconShop~\citep{wu2023iconshoptextguidedvectoricon}               &                                                 & \checkmark                                   & \checkmark     & SVG            \\ 
DLT~\citep{levi2023dltconditionedlayoutgeneration}                    &  \checkmark                                              &                             & $\triangle$           & Layout       \\
DP-TBART~\citep{castellon2023dptbarttransformerbasedautoregressivemodel}                &                                             &  \checkmark                                     &           & Tabular Data \\ 
TabNAT~\citep{zhang2025tabnat}                 & \checkmark                                             &                                       &  & Tabular Data \\ \midrule
\textbf{IPAM (Ours)}            & \boldcheckmark                                             & \boldcheckmark                                   & \boldcheckmark     & \textbf{Domain-Agnostic}         \\ \bottomrule
\end{tabular}
}
}
}
\vspace{-5pt}
\end{table*}


%
Despite the growing importance of hybrid discrete-continuous sequences, conventional approaches face fundamental limitations.
These methods typically discretize continuous values into discrete tokens~\citep{gupta2021layouttransformerlayoutgenerationcompletion, wu2023iconshoptextguidedvectoricon, 2021deepcad, gulati2023tabmtgeneratingtabulardata} or rely on image rasterization~\citep{2022clipdraw, jain2023vectorfusion, 2024svgdreamer}, both fundamentally compromising precision by imposing finite resolution limits~(\cref{fig:intro_circuit}).
This precision sacrifice becomes particularly critical in high-precision domains such as semiconductor circuit design, where even minor positioning errors can lead to complete system failure.
Recently, MAR~\citep{li2024autoregressiveimagegenerationvector} extended autoregressive modeling to purely continuous sequences for image generation, but remains limited to fixed-sized image outputs. 
Meanwhile, a few existing methods~\citep{levi2023dltconditionedlayoutgeneration, zhang2025tabnat} successfully combine discrete and continuous values by preserving their nature, but are limited to fixed schemas~(tabular data) or rely on non-autoregressive diffusion, both of which are inadequate for variable-length sequential generation.

To address these challenges, we propose IPAM (Infinite-Precision Autoregressive Model), a novel autoregressive framework that jointly models discrete and continuous values while supporting variable-length generation~(\cref{tab:scope_comparisons}).
Specifically, discrete values are predicted through traditional categorical prediction, while continuous values are handled via diffusion-based probabilistic models.
Our method provides a unified latent representation of both value types, seamlessly integrating them within an autoregressive structure.
To generate outputs of contextually appropriate lengths, we propose two technical components.
First, we introduce an end-of-sequence (EOS) control mechanism that employs an MLP to dynamically adjust these token logits based on the context.
Second, we incorporate a length regularization term into the loss function, enabling differentiable length control during training.
We demonstrate through our experiments that these components help the model generate sequences closer to ground-truth lengths, thus achieving performance enhancement.

In addition, we introduce ContLayNet, a benchmark specifically designed to evaluate hybrid discrete-continuous sequence generation in real-world engineering applications.
The ContLayNet dataset comprises 334K nano-scale semiconductor layout samples with naturally variable-length sequences, represented as high-precision vectors and collected from real-world sources. 
This benchmark addresses the scarcity of datasets, where precision errors significantly impact functional performance, enabling rigorous evaluation of hybrid generation methods.
To facilitate a comprehensive evaluation, we also propose specialized metrics based on Design Rule Checks (DRCs).

The main contributions of this paper are summarized below: \vspace{-7mm}
\begin{itemize}
    \item We reveal the core limitations of existing deep learning methods in handling hybrid discrete-continuous vector data, especially those relying on discretization, which often compromise precision or structural integrity. \vspace{-2mm}
    
    \item We propose IPAM, a novel autoregressive framework that jointly models discrete and continuous values within variable-length sequences, bridging both kinds of representations without lossy transformations. To ensure contextually appropriate sequence lengths, we introduce an MLP-based EOS logit adjustment mechanism and a length regularization loss that enables differentiable length control during training.
\vspace{-2mm}
    
    \item We construct ContLayNet, a large-scale benchmark comprising high-precision hybrid vector representations of real-world semiconductor layouts, alongside specialized metrics to enable rigorous evaluation of generative models in this domain.
\vspace{-2mm}
    
    \item We demonstrate IPAM's effectiveness across multiple domains---semiconductor layouts, graphic layouts, and text-to-SVG synthesis---outperforming existing discretization-based and fixed-schema methods, particularly in high-precision settings with variable-length hybrid sequences.
    
\end{itemize}



\section{Related Works}
\label{sec:relwork}

Current machine learning approaches for vector representation employ various techniques, but most compromise precision due to inherent discretization errors or finite resolution limits.
Discretization methods convert continuous coordinates into discrete tokens~\citep{gupta2021layouttransformerlayoutgenerationcompletion, wu2023iconshoptextguidedvectoricon, 2021deepcad}, while rasterization-based approaches~\citep{jain2023vectorfusion, 2024svgdreamer} employ differentiable renderers~\citep{2020diffvg} or leverage pretrained vision models~\citep{2022clipdraw, 2023diffsketcher} like CLIP~\citep{radford2021learning}.
Consequently, both paradigms fundamentally restrict optimization by imposing finite resolution or vocabulary constraints, severely hindering their scalability to high-precision domains.
Despite impressive results, these methods face inherent limitations in scalability.
While a few frameworks~\citep{2023svgformer, levi2023dltconditionedlayoutgeneration, zhang2025tabnat} preserve the continuous nature of hybrid representations, they remain domain-specific or bound to fixed schemas, which limits their general applicability to complex, variable-length sequences.
Similarly, recent efforts~\citep{chen2024diffusion, zhao2024pard, li2025layerdag} combining autoregressive architectures with diffusion models are restricted to specific settings---such as discrete graph generation or fixed-length continuous sequences---leaving the joint modeling of hybrid representations with variable-length control largely unaddressed.
To overcome these barriers, we present IPAM, a domain-agnostic framework that models variable-length hybrid sequences through a unified autoregressive architecture that seamlessly integrates both discrete and continuous values.



\section{Preliminaries}
\label{sec:preliminaries}

This section establishes the conceptual foundations necessary for understanding our work.
We define precision formulations and highlight scaling challenges in~\cref{subsec:precision}, and then review autoregressive models from traditional discrete approaches to recent continuous extensions, along with their limitations in~\cref{subsec:autoregressive}.

\subsection{Precision analysis}
\label{subsec:precision}
For each continuous dimension $x \in X$, we define its precision as follows:
\begin{equation}
    P_x = \log_2 \left(\frac{x_\text{max}}{\Delta x} \right),
\end{equation}
where $x_\text{max} = \sup(X)$ is the supremum of the domain, and $\Delta x = \inf(|x_a - x_b|)$ denotes the minimum non-zero distance between two distinct points, $x_a, x_b \in X$.

Hybrid vector representations preserve continuous values, theoretically supporting \textit{infinite} precision~(as $\Delta x \rightarrow 0$).
This property ensures exact positioning and scale invariance, which is critical for high-precision applications---such as semiconductor circuit design---where nanometer-level accuracy is imperative for functional success.

In contrast, discretization imposes a lower bound on $\Delta x$.
For example, a $200 \times 200$ discrete grid with integer quantization~($\Delta x = 1$) yields only $P_x = P_y \approx 7.6$ bits of precision.
To match the precision of continuous representations, the discretized vocabulary size must grow exponentially; achieving $P_x$ bits of resolution requires $2^{P_x}$ unique tokens.

This exponential growth creates a critical trade-off between precision and computation; higher precision demands exponentially more tokens, leading to prohibitive computational costs and training instability due to vocabulary explosion.
We present this limitation in the appendix, where we show how increasing the precision in LayoutTransformer~\citep{gupta2021layouttransformerlayoutgenerationcompletion} leads to significant performance degradation.

\subsection{Autoregressive models}
\label{subsec:autoregressive}

Autoregressive models~\citep{vaswani2017attention, devlin2019bertpretrainingdeepbidirectional, brown2020languagemodelsfewshotlearners} operate on discrete token spaces, modeling sequences through conditional probabilities:
\begin{equation}
	p(x^1, \dots, x^n) = p(x^1) \prod_{i=2}^{n} p(x^i|x^1, \dots, x^{i-1})
\end{equation}
These models predict discrete tokens from a fixed vocabulary via categorical distributions, introducing the precision-computation trade-offs discussed in~\cref{subsec:precision}.

Recently, MAR~\citep{li2024autoregressiveimagegenerationvector} has extended autoregressive modeling to continuous spaces using diffusion processes.
At each step, the autoregressive network produces a conditioning vector $\mathbf{z} \in \mathbb{R}^D$, which guides a compact diffusion network to model the continuous conditional distribution $p(x|\mathbf{z})$.
However, MAR is designed for fixed-size image generation with purely continuous values.
In contrast, IPAM jointly models discrete and continuous components within variable-length sequences, addressing the broader challenge of hybrid vector representations across diverse domains.


\section{IPAM Model}
\label{sec:IPAM}

This section presents our autoregressive generation framework designed to process hybrid vector representations.
%


\vspace{-2mm}
\subsection{Atomic unit representation}
\label{subsec:atomic_unit}

\begin{figure}[t]
\centering 
\includegraphics[width=\linewidth]{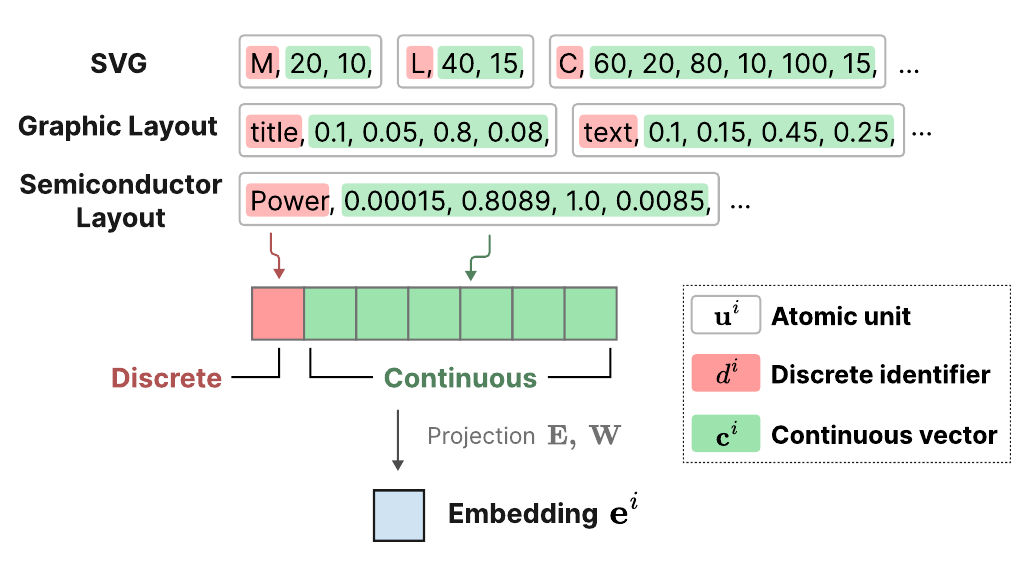}
\vspace{-5mm}
\caption{\small
\textbf{Atomic unit representation.} IPAM processes sequences of atomic units $\mathbf{u}_i$, each consisting of a discrete identifier $d_i$~(red) and a continuous vector $\mathbf{c}_i$~(green).
}
\vspace{-3mm}
\label{fig:IPAM_embedding}
\end{figure}

Many real-world domains, including layouts and SVGs, naturally contain both discrete~(\textit{e.g.}, types and classes) and continuous values~(\textit{e.g.}, coordinates and amounts).
The majority of such hybrid vector representations can be expressed as sequences of \textit{atomic units}, with each unit combining discrete and continuous components~(\cref{fig:IPAM_embedding}).
This paper focuses on domains that can be structured in this manner.

To formalize this representation, consider a sequence of $N$ atomic units
\begin{equation} \label{eq:unit_sequence}
    \mathbf{U}=[\mathbf{u}^1, \mathbf{u}^2, \cdots, \mathbf{u}^N].
\end{equation}
Each atomic unit consists of a discrete identifier $d^i$ and a continuous vector $\mathbf{c}^i$ as
\begin{equation} \label{eq:unit}
\mathbf{u}^i = [d^i, \mathbf{c}^i],
\end{equation}
where $d^i$ takes discrete values in $\{0, \ldots, K-1\}$ and $\mathbf{c}^i = (c_1^i, c_2^i, ..., c_M^i)$ consists of $M$ real-valued scalars.
%


For the example of semiconductor and graphic layouts, an atomic unit is structured as
\begin{equation} \label{eq:layout}
    \mathbf{u}^i=[d^i, x^i, y^i, w^i, h^i]
\end{equation}
where $d^i$ specifies the element type (e.g., power, wiring, and device layers in ContLayNet; or text, title, list, table, and figure in PubLayNet~\citep{zhong2019publaynetlargestdatasetdocument}), and $(x^i, y^i, w^i, h^i)$ denote the position and size.
Similarly, in SVGs, an atomic unit takes the following form:
\begin{equation} \label{eq:svg}
    \mathbf{u}^i=[d^i, x_1^i, y_1^i, \cdots, x_4^i, y_4^i]
\end{equation}
where $d^i$ denotes the path command type~(M, L, C) and the subsequent coordinate pairs represent the positions of the respective control points.

\subsection{Methodology}
\label{subsec:method}
IPAM leverages a unified architecture that processes discrete and continuous components jointly, enabling seamless integration of both data types.

\vspace{-2mm}
\paragraph{Input embedding.}

For an atomic unit $\mathbf{u}^i$ defined in \cref{eq:unit}, we compute its embedding vector $\mathbf{e}^i \in \mathbb{R}^D$ as
\begin{equation} \label{eq:latent_vector}
    \mathbf{e}^i = \text{concat}(\mathbf{E}(\text{onehot}(d^i)), \mathbf{W} \mathbf{c}^i),
\end{equation}
where $\mathbf{E} \in \mathbb{R}^{D_1 \times K}$ is a learned embedding matrix and $\mathbf{W} \in \mathbb{R}^{D_2 \times M}$ is a learned projection matrix, partitioning the full embedding dimension $D = D_1 + D_2$ into two disjoint subspaces.
The embedding vector $\mathbf{e}^i$ thus effectively incorporates information from both discrete and continuous components.


\vspace{-2mm}
\paragraph{Architecture.}

\begin{figure}[t]
\centering 
\includegraphics[width=0.5\textwidth]{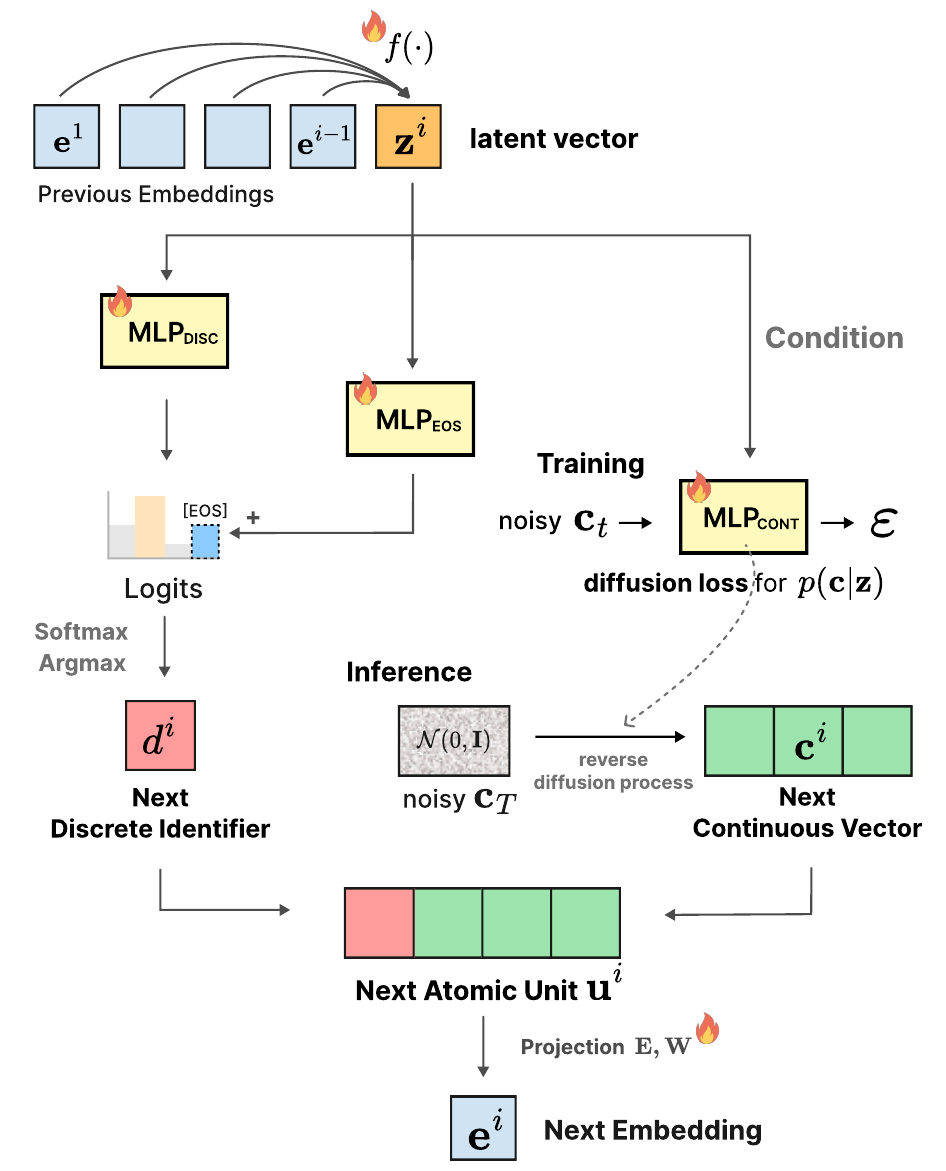}
\caption{\small
\textbf{Overview of IPAM.} IPAM jointly generates discrete and continuous values through the two separate branches.} 
\vspace{-2mm}
\label{fig:IPAM_pipeline}
\end{figure}

The autoregressive network $f(\cdot)$ processes a sequence of embedding vectors $(\mathbf{e}^1, \mathbf{e}^2, \cdots, \mathbf{e}^{i-1})$ to generate a latent vector $\mathbf{z}^i$ at the current step $i$ as
\begin{equation}
    \mathbf{z}^i = f(\mathbf{e}^1, \mathbf{e}^2, \cdots, \mathbf{e}^{i-1}),
\end{equation}
Note that this latent representation $\mathbf{z}^i$ conditions two parallel branches for discrete and continuous generations.
In the discrete branch, $\mathbf{z}^i$ is converted into categorical probabilities using a multi-layer perceptron~(MLP$_\text{DISC}$) followed by a softmax function:
\begin{equation}
    p(d^i|\mathbf{z}^i) = \text{softmax}(\text{MLP}_{\text{DISC}}(\mathbf{z}^i))
\end{equation}
In the continuous branch, a denoising network $\boldsymbol{\varepsilon}_\theta$, conditioned on $\mathbf{z}^i$, iteratively removes noise from corrupted target vectors, enabling sampling from the continuous distribution $p(\mathbf{c}^i|\mathbf{z}^i)$.
Following MAR~\citep{li2024autoregressiveimagegenerationvector}, $\boldsymbol{\varepsilon}_\theta$ is implemented as a small MLP consisting of residual blocks with AdaLN~\citep{peebles2023scalablediffusionmodelstransformers}.

%
%

\vspace{-2mm}
\paragraph{EOS logit adjustment.}
The EOS token signifies a global decision to terminate the sequence, requiring a comprehensive understanding of the entire synthesized structure, which is often underrepresented in the discrete categorical distribution.
To empower the model with such a high-fidelity termination signal, we introduce an EOS-specific adjustment mechanism. 
This allows the model to perform a final structural audit by explicitly referencing the continuous latent state $\mathbf{z}^i$.
The adjustment is implemented as:
%
%
\begin{equation}
    \text{logits}^i[\text{EOS}] \leftarrow \text{logits}^i[\text{EOS}] + \alpha \cdot \text{MLP}_{\text{EOS}}(\mathbf{z}^i)
\end{equation}
where $\alpha$ is a scaling factor and $\text{MLP}_{\text{EOS}}(\cdot)$ is the logit-adjustment function based on the current sequence state.
By incorporating this logit adjustment, the model can make more robust decisions regarding sequence termination, effectively reducing premature or delayed stops.

\vspace{-2mm}
\paragraph{Training.}
The discrete component is trained using Cross-Entropy~(CE) loss:
\begin{equation} \label{eq:loss_disc}
    \mathcal{L}_{d}(\mathbf{z}^i, d^i) = \text{CE}(\text{onehot}(d^i), p(d^i|\mathbf{z}^i))
\end{equation}
The continuous component is trained using a denoising score-matching objective:
\begin{equation} \label{eq:loss_cont}
    \mathcal{L}_{c}(\mathbf{z}^i, \mathbf{c}^i) = \mathbb{E}_{\boldsymbol{\varepsilon}, t}\left[ \lVert \boldsymbol{\varepsilon} - \boldsymbol{\varepsilon}_\theta(\mathbf{c}^i_t | t, \mathbf{z}^i) \rVert^2 \right]
\end{equation}
where $\boldsymbol{\varepsilon} \sim \mathcal{N}(\mathbf{0}, \mathbf{I})$ is Gaussian noise and $\mathbf{c}^i_t = \sqrt{\bar{\alpha}_t}\mathbf{c}^i + \sqrt{1-\bar{\alpha}_t}\boldsymbol{\varepsilon}$ is a noise-corrupted vector at timestep $t$. 
%

We also integrate a \textit{length regularization loss} that encourages generated sequences to align with target lengths during training.
After applying the EOS logit adjustment, the expected sequence length $\ell_{\text{exp}}$ is computed from the adjusted EOS probability, which is given by
\begin{equation}
	\ell_{\text{exp}}(p_{\text{EOS}}) = \sum_{i=1}^N i \cdot p_{\text{EOS}}^i \prod_{j=1}^{i-1}[1 - p_{\text{EOS}}^j] \label{eq:expected_length}
\end{equation}
where $p_{\text{EOS}}^i$ is the probability of the EOS token at position $i$.
The length regularization loss $\mathcal{L}_\ell$ is then defined by
\begin{equation}
\mathcal{L}_{\ell} = (\ell_{\text{exp}}(p_{\text{EOS}}) - \ell_{\text{target}})^2. \label{eq:length_loss}
\end{equation}
%
%
This formulation is fully differentiable, enabling end-to-end optimization of sequence lengths alongside content generation.
The total training objective combines the two loss components with the regularization term, which  is given by
\begin{equation}
    \mathcal{L}_\text{total} = \mathcal{L}_{d} + \lambda_1\cdot\mathcal{L}_{c} + \lambda_2\cdot\mathcal{L}_{\ell}
\end{equation}
where $\lambda_1$ and $\lambda_2$ balance the three loss terms.

\vspace{-2mm}
\paragraph{Generation process.}
For discrete identifiers, samples are drawn from $p(d|\mathbf{z})$ after the EOS adjustment.
For continuous vectors, samples are generated through a reverse diffusion process:
\begin{equation}
    \mathbf{c}_{t-1} = \frac{1}{\sqrt{\alpha_t}} \left( \mathbf{c}_t - {1 - \alpha_t \over \sqrt{1-\bar{\alpha}_t}} \boldsymbol{\varepsilon}_\theta(\mathbf{c}_t|t, \mathbf{z})  \right) + \sigma_t \boldsymbol{\xi}
\end{equation}
where $\boldsymbol{\xi} \sim \mathcal{N}(\mathbf{0}, \mathbf{I})$ is a noise vector, and $\sigma_t$ controls the noise level.
Starting from an initial noise vector $\mathbf{c}_T \sim \mathcal{N}(\mathbf{0}, \mathbf{I})$, this iterative process generates samples that follow the target distribution $p(\mathbf{c}|\mathbf{z})$.
By generating each component in its natural form, our approach preserves high precision while maintaining the inherent structure of hybrid vector representations.

\begin{figure*}[t]
    \centering
    \includegraphics[width=0.95\linewidth]{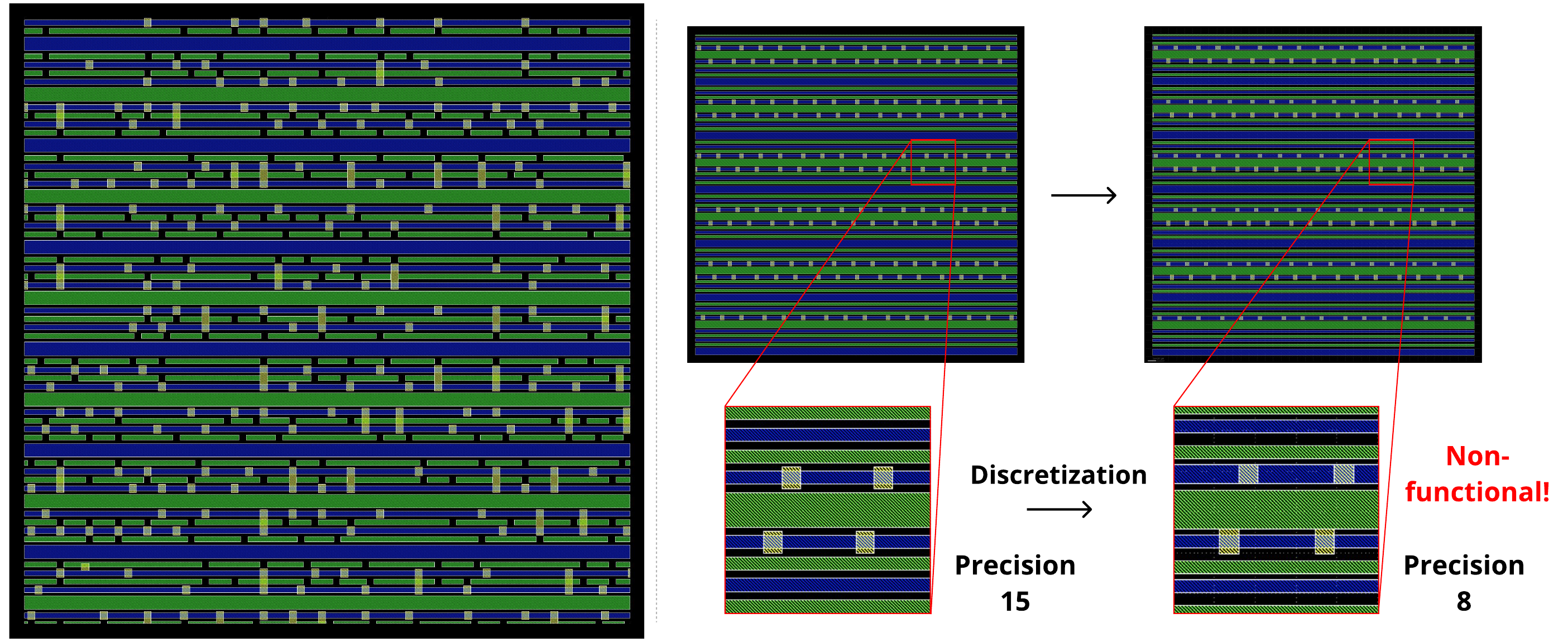}
    \caption{\textbf{The ContLayNet dataset.} Left: Example visualization of a sample. Right: Comparison showing functional failure when ContLayNet samples get discretized to low precision.}
    \vspace{-3mm}
    \label{fig:contlaynet}
\end{figure*}


\vspace{-2pt}
\section{ContLayNet Benchmark}
\label{sec:contlaynet}

This section provides details on the dataset and evaluation metrics in the ContLayNet benchmark.

\subsection{Dataset}
\label{subsec:dataset}


The ContLayNet dataset consists of 334,330 semiconductor layout samples, with original continuous values converted to an effective 15-bit precision~($40,000 \times 40,000$ integer grid)\footnote{Refer to our project page https://github.com/yxxshin/IPAM.}.
Each sample contains three primary layers---Power, Wiring, and Device---visualized in blue, green, and yellow, respectively, in~\cref{fig:contlaynet}.
Cells specify each layer's type, position, and size.  
Power~(Layer 515), Wiring~(Layer 644), and Device~(Layer 1457) layers contain an average of 30, 116, and 178 elements, respectively, with standard deviations of 10, 92, and 75.
The dataset exhibits substantial structural complexity, with each sample including an average of 323 layers.
The original continuous values are discretized and rounded to a $40,000 \times 40,000$ integer grid with substantial structural complexity, each sample averaging 323 layers.

\cref{fig:contlaynet} illustrates how precision reduction leads to functional failures: low-precision discretization disrupts critical geometric alignments between layers, ultimately rendering circuits non-functional.
This property makes ContLayNet particularly well-suited to our research objectives; it offers a real-world dataset in which precision loss directly results in measurable performance degradation.

\subsection{Evaluation Metrics}
\label{subsec:eval_metric}

Evaluating semiconductor layouts presents unique challenges, as traditional image-based metrics like FID~\citep{heusel2018ganstrainedtimescaleupdate} and CLIP Score~\citep{radford2021learning} fail to capture the critical internal semantics required for functional quality assessment.
To address this gap, we introduce specialized evaluation metrics based on Design Rule Checks~(DRCs), which assess whether generated layouts satisfy the essential constraints of real-world circuit designs.
These metrics quantitatively measure the model's generation performance in preserving critical design rules and geometric precision.
%

We define four fundamental design rules: the Power Delivery Constraint~(PDC), Circuit Linkage Constraint~(CLC), Horizontal Spacing Constraint~(HSC), and Vertical Spacing Constraint~(VSC).
%
CLC measures the overall functionality of the layout, while PDC, HSC, and VSC evaluate whether device layers are placed correctly.
To quantify performance, we compute the non-functional~(overlapped-circuit) area for CLC and violation rates by counting layers that breach the other three rules.
Detailed explanations of these metrics are provided in~\cref{subsec:drc_details}.
\vspace{-5pt}
\begin{enumerate}
    \item \textbf{Power Delivery Constraint:} Layer 1457 must reside above Layer 515 to ensure adequate power delivery. \vspace{-1mm}
    \item \textbf{Circuit Linkage Constraint:} Layer 644 and Layer 515 must not overlap each other to maintain the overall functionality of the circuit. \vspace{-1mm}
    \item \textbf{Horizontal Spacing Constraint:} Instances of Layer 1457 aligned vertically must maintain a minimum horizontal separation of $W$. \vspace{-1mm}
    \item \textbf{Vertical Spacing Constraint:} Instances of Layer 1457 aligned horizontally must maintain a minimum vertical separation of $H$.
\end{enumerate}
\vspace{-2mm}
%
We normalize each metric to $[0, 1]$ by dividing the number of violating components by the maximum possible violations.
For the Horizontal and Vertical Spacing Constraints, which measure the minimum pairwise distance between Device~(Layer 1457) layers, we introduce a penalty for samples with fewer than a predetermined threshold $\eta$ of Device layers.
This prevents models from artificially improving spacing scores by simply omitting critical components, as layouts with fewer Device layers can trivially satisfy large minimum-distance requirements.
For HSC and VSC, $\eta$, $W$, and $H$ are set to 240, 1200, and 1000, respectively, during the experiments.


\vspace{-2mm}
\section{Experiments}
\label{sec:Exp}

This section presents the experimental setup and evaluates IPAM compared to existing methods across three domains: semiconductor layouts, graphic layouts, and SVGs.
Note that comprehensive implementation and experimental details omitted from the main text are provided in Appendix.

\vspace{-2mm}
\subsection{ContLayNet: semiconductor layout generation}
\label{subsec:gds_exp}

\begin{table*}[t!]
\caption{\textbf{Comparison on the ContLayNet Benchmark.} IPAM outperforms baseline methods across every metric. Both EOS logit adjustment and length regularization components contribute to quality. Bold indicates best performance.}
\label{tab:high_prec}
\vspace{-3mm}
\begin{center}
\setlength{\tabcolsep}{11pt}
\scalebox{0.85}{
\begin{tabular}{lc|cccc|cccc}
\toprule
  &    & \multicolumn{4}{c|}{Completion~(given 50 layers)}                  & \multicolumn{4}{c}{Completion~(given 100 layers)}  \\
Model & Precision & CLC $\downarrow$  & PDC $\downarrow$ & HSC $\downarrow$ & \multicolumn{1}{c|}{VSC $\downarrow$} & CLC $\downarrow$  & PDC $\downarrow$ & HSC $\downarrow$  & VSC $\downarrow$    \\ \midrule
LT & 18   & 0.397 & 0.149 & 0.053 & \multicolumn{1}{c|}{0.058} & 0.152 & 0.144 & 0.051 & 0.054    \\ 
DLT & $\infty$  & 0.227 & 0.125 & 0.065  & \multicolumn{1}{c|}{0.056} & 0.144 & 0.119 & 0.065 & 0.052    \\ \midrule
IPAM & $\infty$ & $\textbf{0.088}$ & $\textbf{0.068}$ & $\textbf{0.027}$   & \multicolumn{1}{c|}{$\textbf{0.023}$} &$\textbf{0.046}$ & $\textbf{0.058}$ & $\textbf{0.024}$ & $\textbf{0.025}$    \\ 
 \ \ - w/o $\text{MLP}_{\text{EOS}}$ & & 0.101 & 0.078 & 0.028 &  \multicolumn{1}{c|}{0.034} & 0.050 & 0.092 & 0.027 & 0.033 \\
 \ \ - w/o $\mathcal{L}_{\ell}$ & & 0.107 & 0.076 & 0.032 &  \multicolumn{1}{c|}{0.038} & 0.067 & 0.068 & 0.031 & 0.032
 \\
 \ \ - w/o $\text{MLP}_{\text{EOS}}$ and $\mathcal{L}_{\ell}$ & & 0.101 & 0.113 & 0.030   & \multicolumn{1}{c|}{0.029} & 0.056 & 0.139 & 0.031 & 0.034 \\
\midrule
Real  &  $\infty$     & 0.0 \ \ \ \ & 0.0 \ \ \ \ & 0.0 \ \ \ \ & \multicolumn{1}{c|}{0.0 \ \ \ \ } & 0.0 \ \ \ \ & 0.0 \ \ \ \ & 0.0 \ \ \ \ & 0.0 \ \ \ \     \\ \bottomrule
\end{tabular}
}
\end{center}
\end{table*}

\begin{figure*}[t!]
    \begin{center}
    \vspace{-2mm}
    \includegraphics[width=0.95\linewidth]{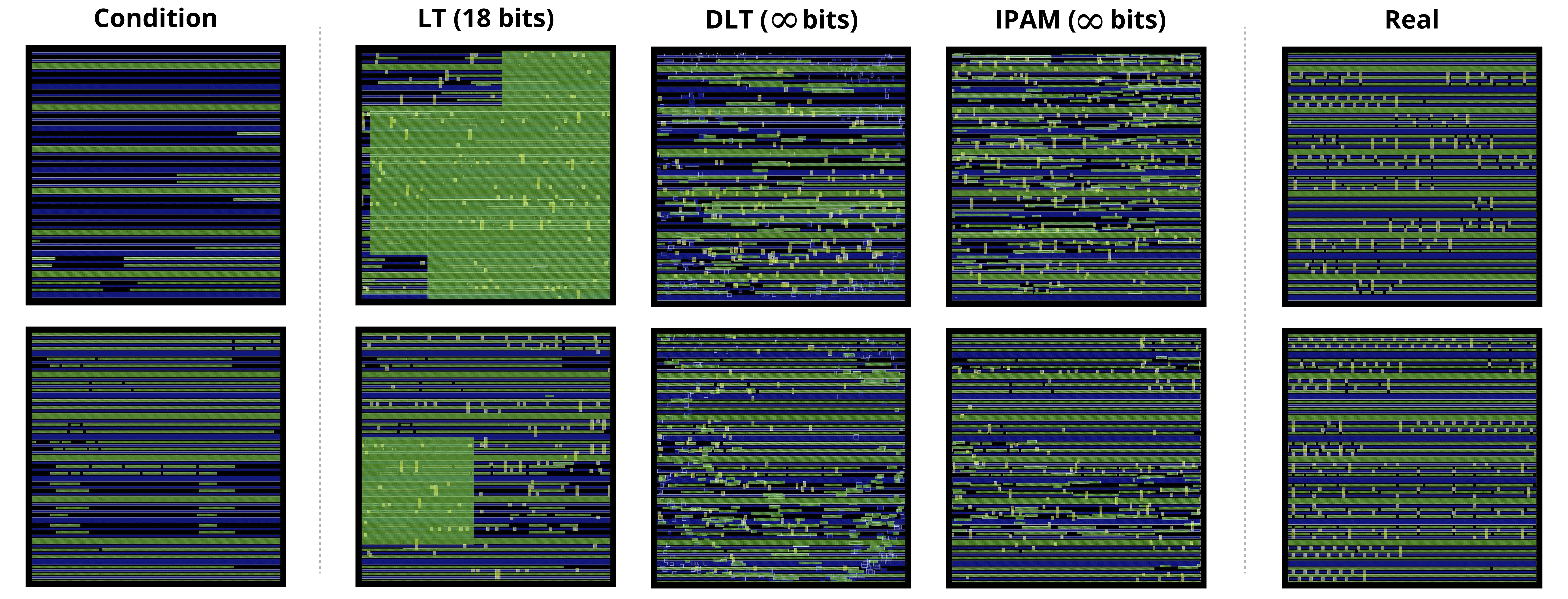}
    \vspace{-1mm}
    \caption{\textbf{Qualitative results on ContLayNet in completion task given 50 / 100 layers.} 
    IPAM illustrates clearly superior performance.}
    \vspace{-2mm}
    \label{fig:gds_qual}
    \end{center}
\vspace{-4mm}
\end{figure*}

\paragraph{Baselines.}
We compare IPAM with two representative baselines: LayoutTransformer~(LT)~\citep{gupta2021layouttransformerlayoutgenerationcompletion}, an autoregressive model tailored for sequential data generation but constrained by token discretization, and DLT~\citep{levi2023dltconditionedlayoutgeneration}, a non-autoregressive diffusion model designed for graphic layout generation that jointly models discrete and continuous values.
For a fair comparison, we evaluate LT using 18-bit precision (the maximum precision achievable with a batch size of 2 on a single NVIDIA A6000 GPU), while retaining the original hybrid vector representations for DLT and IPAM.

\vspace{-2mm}
\paragraph{Implementation details.}
IPAM employs a Transformer decoder architecture with a hidden dimension of 1024, comprising 32 decoder layers and 16 attention heads.
$\text{MLP}_{\text{CONT}}$ consists of three blocks with 1024 channels each, while $\text{MLP}_{\text{DISC}}$ and $\text{MLP}_{\text{EOS}}$ are both two-layer MLPs with hidden layers twice the width of their input dimensions.
We set $\lambda_1 = 100$, $\lambda_2 = 0.1$, and $\alpha = 0.1$.
%
%
%

\vspace{-2mm}
\paragraph{Results.}
\cref{tab:high_prec} presents the performance of IPAM on ContLayNet.
Across both completion tasks involving 50 and 100 layers, IPAM consistently outperforms all baselines across the four metrics: CLC, PDC, HSC, and VSC.
These results demonstrate the clear superiority of our proposed framework.
\cref{fig:gds_qual} illustrates the qualitative results of IPAM. 
LT exhibits oversized layers that overlap large portions of the layout, often obscuring underlying structures.
DLT tends to generate numerous noisy elements due to its limited length control capabilities~(\cref{fig:appendix_magnified_DLT}).
In contrast, IPAM produces more balanced and well-distributed layers than both LT and DLT, showing strong alignment with real samples while leaving room for further optimization.

\vspace{-2mm}
\paragraph{Ablation studies.}
\cref{tab:high_prec} also includes ablation studies for our two key components: the EOS logit adjustment ($\text{MLP}_{\text{EOS}}$) and the length regularization term ($\mathcal{L}_\ell$).
The results demonstrate that both components significantly contribute to performance, with their combination achieving the best results across every metric.
In terms of length control, our complete IPAM framework achieves a near-zero mean error ($\mu = 0.15$) with reduced variance ($\sigma = 146.13$), whereas the ablated version exhibits both a noticeable bias ($\mu = 38.91$) and a higher variance ($\sigma = 154.45$) as illustrated in \cref{fig:ablation_gaussian} in Appendix.

%
%

\vspace{-1mm}
\subsection{Graphic layout generation}
\label{subsec:lg_exp}

\vspace{-1mm}
\paragraph{Experimental setup.}
\label{subsubsec:lg_setup}
We evaluate our model on \textit{PubLayNet}~\citep{zhong2019publaynetlargestdatasetdocument} and \textit{Rico}~\citep{Deka:2017:Rico}, which contain 330K scientific document layouts and 91K mobile UI layouts, respectively.
%
For evaluation, we use three standard metrics: Fr\'echet Inception Distance~(FID)~\citep{heusel2018ganstrainedtimescaleupdate}, Overlap~\citep{li2019layoutgangeneratinggraphiclayouts}, and Alignment score~\citep{lee2020neuraldesignnetworkgraphic}.
We again compare our approach against LT~\citep{gupta2021layouttransformerlayoutgenerationcompletion} with 18-bit precision, and DLT~\citep{levi2023dltconditionedlayoutgeneration}.
%
%
%

\begin{figure*}[t!]
    \begin{center}
    \vspace{-2mm}
    \includegraphics[width=0.95\linewidth]{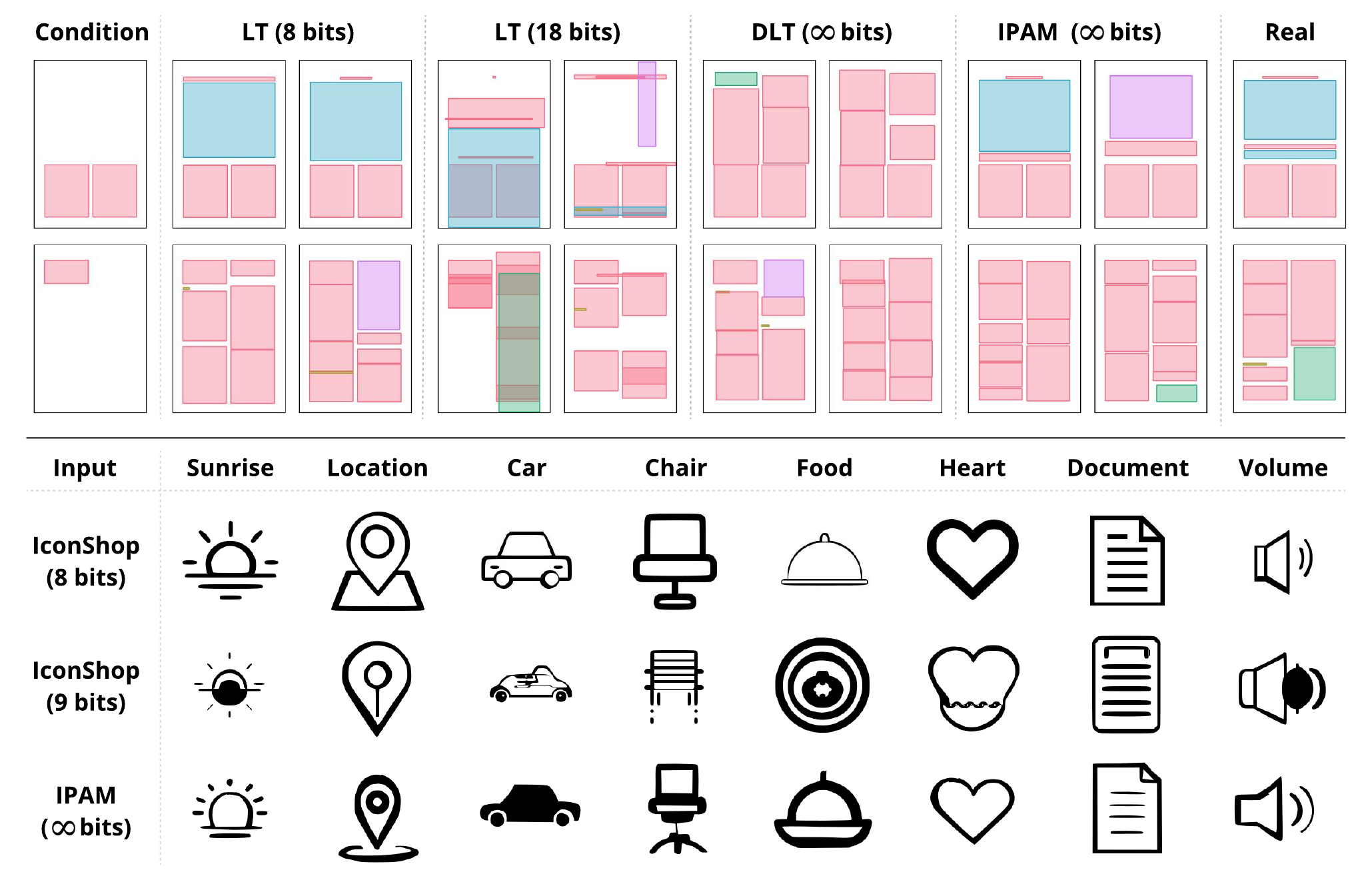}
    \vspace{-2mm}
    \caption{\textbf{Top: Layout Generation on PubLayNet. Bottom: Text-to-SVG Generation on FIGR-8-SVG.} 
    IPAM achieves superior performance in high-precision settings while maintaining quality comparable to low-precision settings across diverse domains.}
    \vspace{-3mm}
     \label{fig:qual_others}
    \end{center}
\end{figure*}

\vspace{-3mm}
\paragraph{Results.}
On both the PubLayNet and Rico datasets, IPAM outperforms LT~(with 18-bit precision) and DLT across all tasks, including completion and unconditioned generation~(\cref{tab:high_prec_layout}).
Specifically, IPAM achieves the best FID, Overlap, and Alignment scores, demonstrating its superiority in complex layouts where baseline models face significant challenges.
Ablation studies on these datasets confirm the effectiveness of both the EOS logit adjustment and length regularization components, with consistent improvements observed across all metrics.
%
\cref{fig:qual_others} illustrates the qualitative results of IPAM, which achieves generation quality comparable to low-precision LT while clearly outperforming its high-precision counterpart and DLT.
We provide additional qualitative results, including unconditioned generation and experiments on the Rico dataset, in Appendix.



\begin{table*}[t!]
\caption{\textbf{Comparison on high-precision graphic layout generation.} IPAM consistently outperforms baselines across all metrics. Metrics: \textit{FID}~(lower is better), \textit{Overlap} and \textit{Alignment} scores~(closer to real is better, $\times 100$ for clarity).}
\label{tab:high_prec_layout}
\vspace{-2mm}
\begin{center}
\setlength{\tabcolsep}{4pt}
\scalebox{0.85}{
\begin{tabular}{lc|cccccc|cccccc}
\toprule
&  & \multicolumn{6}{c|}{\textbf{PubLayNet}}   & \multicolumn{6}{c}{\textbf{Rico}}    \\ \midrule
&      & \multicolumn{3}{c|}{Completion}                  & \multicolumn{3}{c|}{Un-Gen} & \multicolumn{3}{c|}{Completion}                 & \multicolumn{3}{c}{Un-Gen} \\ 
Model & Precision & FID      & Overlap & \multicolumn{1}{c|}{Align}  & FID         & Overlap   & Align    & FID     & Overlap & \multicolumn{1}{c|}{Align}  & FID        & Overlap   & Align    \\ \midrule
LT & 18   & 186.61 & 64.16 & \multicolumn{1}{c|}{1.06} & 198.76    & 59.97   & 1.91   & 24.32 & 45.95 & \multicolumn{1}{c|}{0.35} & 28.29   & 50.84   & 0.41   \\
DLT & $\infty$  & \ \ 35.80    & \ \ 9.65    & \multicolumn{1}{c|}{0.84}   & \ \ 44.00       & \ \ 8.64      & 0.71     & 20.92   & 28.11  & \multicolumn{1}{c|}{0.63}   & 38.97      & 38.87     & 0.39     \\ \midrule
IPAM & $\infty$ & \ \ \ \ \textbf{4.58}    & \ \ \textbf{4.59}    & \multicolumn{1}{c|}{\textbf{0.17}}   & \ \ \textbf{12.36}       & \ \ \textbf{4.00}      & \textbf{0.20}     & \ \ 9.77    & \textbf{33.11}   & \multicolumn{1}{c|}{\textbf{0.25}}   & \textbf{28.19}      & \textbf{31.90}     & \textbf{0.30}    \\
\ \ - w/o $\text{MLP}_{\text{EOS}}$ & & \ \ \ \ 5.27    & \ \ 5.08    & \multicolumn{1}{c|}{\textbf{0.17}}   & \ \ 12.75       & \ \ 4.79      & 0.21     & 10.16    & 32.05   & \multicolumn{1}{c|}{0.26}   & 30.40      & 31.63     & 0.32    \\
\ \ - w/o $\mathcal{L}_\ell$ & &  \ \ \ \ 5.05    & \ \ 4.65    & \multicolumn{1}{c|}{0.18}   & \ \ 12.51       & \ \ 4.67      & 0.21     & \ \ \textbf{9.63}    & 32.07   & \multicolumn{1}{c|}{0.25}   & 28.32      & 30.75     & 0.35    \\ 
\ \ - w/o $\text{MLP}_{\text{EOS}}$ and $\mathcal{L}_{\ell}$  & & \ \ \ \ 5.07    & \ \ 5.25    & \multicolumn{1}{c|}{0.18}   & \ \ 12.90       & \ \ 5.00      & 0.21     & 10.76    & 31.62   & \multicolumn{1}{c|}{0.29}   & 29.29      & 29.74     & 0.36    \\ \midrule
Real     & $\infty$   & --        & \ \ 0.22    & \multicolumn{1}{c|}{0.03}   & --           & \ \ 0.22      & 0.03     & --       & 32.92   & \multicolumn{1}{c|}{0.17}   & --          & 32.92     & 0.17     \\ \bottomrule
\end{tabular}
}
\end{center}
\end{table*}

\begin{table}[t]
\centering
\setlength{\tabcolsep}{11pt}
\caption{\footnotesize
\textbf{Text-to-SVG results.} IconShop performs adequately with 8-bit precision but fails to scale at higher precisions, whereas IPAM achieves high-quality outputs via infinite precision. 
}\label{tab:svg_results}
\vspace{-1mm}
\scalebox{0.85}{
\begin{tabular}{lc|cc}
        \toprule
        Model & Precision & \ FID $\downarrow$ & \ CLIP $\uparrow$ \\
        \midrule
         & 8 & \ 37.35  & \ 22.13 \\
        IconShop  & 9 &  \ 95.96 & \ 18.07 \\
         & 10 & \ Fail & \ Fail \\ \midrule
        IPAM  & $\infty$ &  \ 48.73  & \ 20.05 \\
        \ \ - w/o $\text{MLP}_{\text{EOS}}$ and $\mathcal{L}_{\ell}$  & $\infty$ &  \ 58.22  & \ 19.90 \\
        \bottomrule
    \end{tabular}
}
\vspace{-4mm}
\end{table}

\vspace{-1mm}
\subsection{Text-to-SVG generation}
\label{subsec:svg_exp}

\vspace{-1mm}
\paragraph{Experimental setup.}
We compare IPAM against IconShop~\citep{wu2023iconshoptextguidedvectoricon}---the only other text-to-SVG method operating at the control-point level---in both high- and low-precision settings to evaluate its capability in preserving the continuous nature of text-to-SVG generation.
To this end, we use the \textit{FIGR-8-SVG} dataset~\citep{clouatre2019figr}, which contains 1.5M SVG-formatted black-and-white icons, following IconShop's preprocessing protocol.
We employ two standard metrics: FID~\citep{heusel2018ganstrainedtimescaleupdate} and CLIP score~\citep{radford2021learning}.
%

\vspace{-2mm}
\paragraph{Results.}
\cref{tab:svg_results} and \cref{fig:qual_others} illustrate IPAM's superior performance in high-precision settings.
While IconShop produces visible results at 8-bit precision but degrades at 9 bits and fails completely beyond 10 bits, IPAM robustly generates continuous vector outputs with its inherent infinite precision, maintaining generation quality comparable to IconShop's best outputs.
Ablation studies further confirm that both key components, $\text{MLP}_{\text{EOS}}$ and $\mathcal{L}_\ell$, consistently contribute to the performance gains.


\section{Conclusion}
\label{sec:conclusion}

This paper introduces IPAM, a novel autoregressive framework that jointly models discrete and continuous components within variable-length sequences, addressing the fundamental limitations of discretization-based approaches.
Our key technical contributions include an EOS control mechanism with MLP-based logit adjustment and a length regularization loss, both of which are critical to achieving performance gain.
To evaluate our method, we introduce ContLayNet, a large-scale benchmark comprising 334K high-precision semiconductor layouts accompanied by specialized design rule checking~(DRC) metrics, thereby bridging a critical gap in datasets where precision errors directly compromise functional performance.
Comprehensive experiments across semiconductor layouts, graphic layouts, and text-to-SVG synthesis demonstrate IPAM’s superior capability in high-precision scenarios, consistently outperforming existing discretization-based and fixed-schema baselines.
We anticipate that our framework will advance hybrid vector
representation learning with ContLayNet serving as a standard benchmark for high-precision engineering domains.
%

%




\clearpage
\section*{Impact Statement}
This work aims to advance the broader field of machine learning by developing robust hybrid generative models.
While our framework focuses on technical domains such as semiconductor layouts, graphic layouts, and vector graphics, we acknowledge that such generative models must be developed and deployed with careful consideration of their dual-use potential and broader societal impacts.
Furthermore, while IPAM consistently outperforms existing baselines, all evaluated approaches, including ours, still struggle with the inherent complexity of the ContLayNet benchmark.
This is particularly evident in longer sequences, where error propagation becomes more pronounced, highlighting both the difficulty of high-precision layout generation and promising opportunities for future research.

\section*{Acknowledgment}
This work was partly supported by Samsung Electronics Co., Ltd. and the Institute of Information \& Communications Technology Planning \& Evaluation (IITP) grants [RS-2022-II220959 (No.2022-0-00959), (Part 2)
Few-Shot Learning of Causal Inference in Vision and Language for Decision Making; RS-2026-25526850, High-Efficiency Neural Networks for Artificial General Intelligence (HERMES-Net); No.RS-2021-II211343, Artificial Intelligence Graduate School Program (Seoul National University)] funded by the Korea government (MSIT).

\bibliography{icml2026_agdc}
\bibliographystyle{icml2026}

\newpage
\appendix
\onecolumn

\appendix


\renewcommand{\thefigure}{\Alph{figure}}
\renewcommand{\thetable}{\Alph{table}}

\renewcommand{\theHfigure}{appendix.\Alph{figure}}
\renewcommand{\theHtable}{appendix.\Alph{table}}

\setcounter{figure}{0}
\setcounter{table}{0}


\section{ContLayNet: Semiconductor Layout Generation}
\label{sec:app_contlaynet}

\subsection{Detailed Statistics of ContLayNet dataset}
The original ContLayNet dataset contains 334,330 nanometer-scale semiconductor circuit layout samples in OASIS format.
These files are converted to text files with headers and cells for model training.
The cells specify each layer's type, location, and size.  
Power~(written as Layer 515), Wiring~(written as Layer 644), and Device~(written as Layer 1457) layers contain an average of 30, 116, and 178 elements, respectively, with standard deviations of 10, 92, and 75.
Original~(continuous) values are divided and rounded-off to a $40,000 \times 40,000$ integer grid with substantial structural complexity, each sample averaging 323 layers.

\subsection{Detailed Description of DRC}
\label{subsec:drc_details}
The design rule check calculations are detailed below:

\begin{enumerate}
    \item \textbf{Power Delivery Constraint:} Layer 1457 must reside above Layer 515 to ensure adequate power delivery.
    \begin{equation}
    \forall i: \quad \text{Area}\left(\mathrm{L}_{515} \cap \mathrm{L}_{1457}^{(i)}\right) > 0
    \end{equation}

    \item \textbf{Circuit Linkage Constraint:} Layer 644 and Layer 515 must not overlap each other to maintain overall functionality of the circuit.
    \begin{equation}
        \forall i \neq j: \quad  \text{Area}\biggl( \left(\mathrm{L}_{515} \cap \mathrm{L}_{644}\right) \cup  \left(\mathrm{L}_{515}^{(i)} \cap \mathrm{L}_{515}^{(j)}\right) \cup \left(\mathrm{L}_{644}^{(i)} \cap \mathrm{L}_{644}^{(j)}\right) \biggr) = 0
    \end{equation}

    \item \textbf{Horizontal Spacing Constraint:} Instances of Layer 1457 aligned vertically must maintain a minimum horizontal separation of $W$.
    \begin{equation}
        \forall i \neq j, \text{ if } |\mathrm{L}_{1457,y}^{(i)} - \mathrm{L}_{1457,y}^{(j)}| < \eta:  \qquad |\mathrm{L}_{1457,x}^{(i)} - \mathrm{L}_{1457,x}^{(j)}| \ge W
    \end{equation}

    \item \textbf{Vertical Spacing Constraint:} Instances of Layer 1457 aligned horizontally must maintain a minimum vertical separation of $H$.
    \begin{equation}
    \forall i \neq j, \text{ if } |\mathrm{L}_{1457,x}^{(i)} - \mathrm{L}_{1457,x}^{(j)}| < \eta: \qquad |\mathrm{L}_{1457,y}^{(i)} - \mathrm{L}_{1457,y}^{(j)}| \ge H
    \end{equation}

\end{enumerate}

Here, $\text{Area}$ and Layer (L) represent the computational areas and layer components, respectively, from the ContLayNet dataset at a real-world scale.
We normalize each metric to $[0, 1]$ by dividing the number of violating components by the maximum possible violations.
For Horizontal and Vertical Spacing Constraints, which measure the minimum pairwise distance between Device~(Layer 1457) layers, we add a penalty for samples with fewer than a given threshold number of Device layers.
This prevents models from artificially improving spacing scores by simply omitting critical components, as layouts with fewer Device layers can trivially satisfy large minimum-distance requirements.
For HSC and VSC, $\eta, W, H$ are set as 240, 1200, and 1000 during the experiments. 

\subsection{Experiment Details}
\label{subsec:supp_A3}
\paragraph{Implementation details.} 
IPAM utilizes a Transformer decoder architecture with 32 layers, 16 attention heads, and a hidden dimension of 1024.
For the denoising process, we employ an MLP comprising three blocks, each with 1024 channels, following the diffusion approach of~\citep{nichol2021improveddenoisingdiffusionprobabilistic}.
During training, each latent vector $\mathbf{z}$ is processed through the denoising MLP, sampling the timestep $t$ 30 times each.
$\text{MLP}_{\text{CONT}}$ consists of three blocks with 1024 channels each, while $\text{MLP}_{\text{DISC}}$ and $\text{MLP}_{\text{EOS}}$ are both two-layer MLPs with hidden layers twice the width of their input dimensions.
We set $\lambda_1=100$, $\lambda_2=0.1$ and $\alpha=0.1$.
The model was trained for 10 days on a single NVIDIA A6000 GPU with a learning rate of $7.5\times10^{-5}$.
Since the ContLayNet has been newly introduced, further optimization and refinement may yield improved performance.

\subsection{Inference costs and Acceleration}

\cref{tab:app_inference} presents comprehensive timing comparisons for the completion task with 50 layers, sampling 100 samples with a batch size of 5 on a single RTX 4090 GPU. 
Throughout this paper, IPAM uses Improved DDPM~\citep{nichol2021improveddenoisingdiffusionprobabilistic} with 100 diffusion steps for all reported results unless otherwise specified. 
To address inference speed concerns, we evaluate sampling acceleration by reducing diffusion steps. 
Improved DDPM's learned variance formulation enables effective sampling with fewer steps using uniform stride sequences. 
We also evaluate DDIM~\citep{song2022denoisingdiffusionimplicitmodels} as an alternative sampler, which uses a deterministic non-Markovian process that also allows timestep skipping. 
We also compare against baseline methods LT and DLT.

\begin{table}[h]
\centering
\vspace{-8pt}
\setlength{\tabcolsep}{12pt}
{
\caption{\textbf{Inference time and generation quality comparison across methods.} IPAM with Improved DDPM at 50 steps achieves comparable speed to LT while substantially outperforming on all DRC metrics. IPAM demonstrates favorable speed-quality trade-offs compared to LT and DLT across different sampling configurations. Bold indicates best performance.}
\label{tab:app_inference}
\vspace{10pt}
\scalebox{0.85}{
\begin{tabular}{llcccccc}
\toprule
Model & Sampling Method & steps & Time (s) & CLC $\downarrow$ & PDC $\downarrow$ & HSC $\downarrow$ & VSC $\downarrow$ \\ \midrule
LT    & -               & -          & \ \ 766.12   & 0.375            & 0.148            & 0.049            & 0.049            \\ \midrule
DLT   & -               & -          & \ \ \ \ \ \ 7.64     & 0.202            & 0.179            & 0.070            & 0.053            \\ \midrule
IPAM  & Improved DDPM   & 100        & 1489.13  & \textbf{0.084}   & \textbf{0.067}   & \textbf{0.026}   & \textbf{0.022}   \\ 
      &                 & \ \ 50         & \ \ 876.78   & 0.098            & 0.082            & 0.034            & 0.023            \\
      &                 & \ \ 20         & \ \ 616.88   & 0.114            & 0.090            & 0.035            & 0.029            \\ 
      &                 & \ \ 10         & \ \ 579.97   & 0.179            & 0.088            & 0.097            & 0.057            \\ \midrule
      & DDIM            & 100        & 1800.44  & 0.090            & 0.085            & 0.032            & 0.027            \\
      &                 & \ \ 50         & 1205.35  & 0.107            & 0.085            & 0.032            & 0.031            \\
      &                 & \ \ 20         & \ \ 888.95   & 0.139            & 0.117            & 0.056            & 0.045            \\ 
      &                 & \ \ 10         & \ \ 721.95   & 0.210            & 0.127            & 0.123            & 0.080            \\ \bottomrule
\end{tabular}
}}
\end{table}

\cref{tab:app_inference} demonstrates that IPAM achieves practical inference speeds while delivering superior generation quality. With Improved DDPM sampling at 50 steps, IPAM runs at a comparable speed to LT while substantially outperforming across all DRC metrics. Further acceleration is possible by reducing diffusion steps, though with some performance degradation: at 20 steps, IPAM becomes faster than LT while maintaining quality advantages. We also evaluated DDIM as an alternative sampler, but it showed both slower inference times and lower quality compared to Improved DDPM at equivalent step counts. While DLT achieves the fastest inference through its non-autoregressive architecture, its non-autoregressive nature makes it difficult to control sequence length, resulting in many noisy elements, particularly on variable-length datasets like ContLayNet~(\cref{fig:appendix_magnified_DLT}). These results demonstrate that users can reduce inference time at the cost of generation quality based on their preference, while still maintaining substantially better functional correctness than discretization baselines even at reduced step counts.

\subsection{Additional Results.}

\begin{figure}[t]
\centering 
\includegraphics[width=0.55\linewidth]{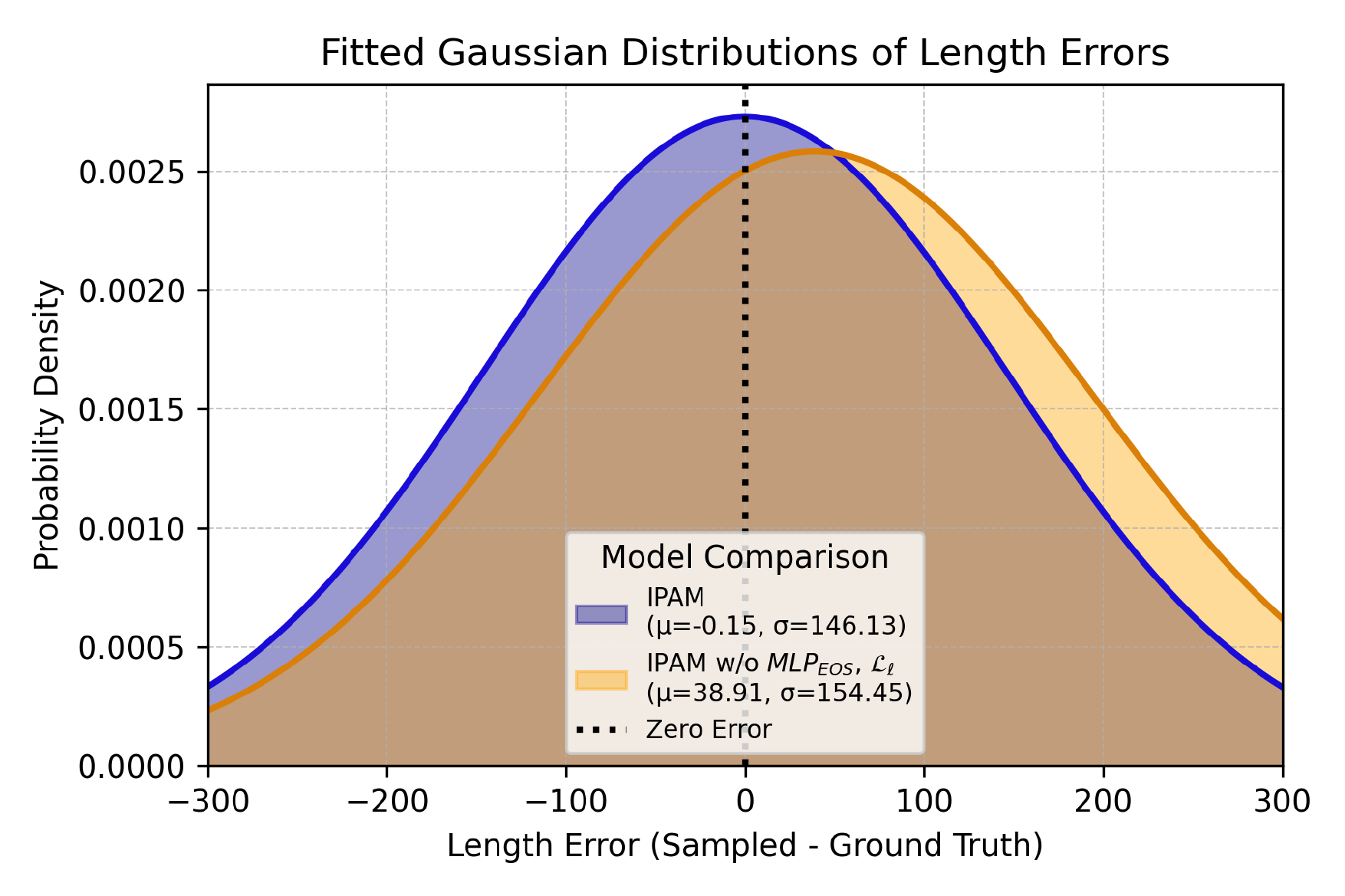}
\caption{\small
\textbf{Length error distribution comparisons.} 
Fitted Gaussian distributions of length errors illustrate the impact of EOS logit adjustment and length regularization on length control accuracy.
}
\label{fig:ablation_gaussian}
\end{figure}

\begin{figure*}[t!]
    \begin{center}
    \includegraphics[width=\linewidth]{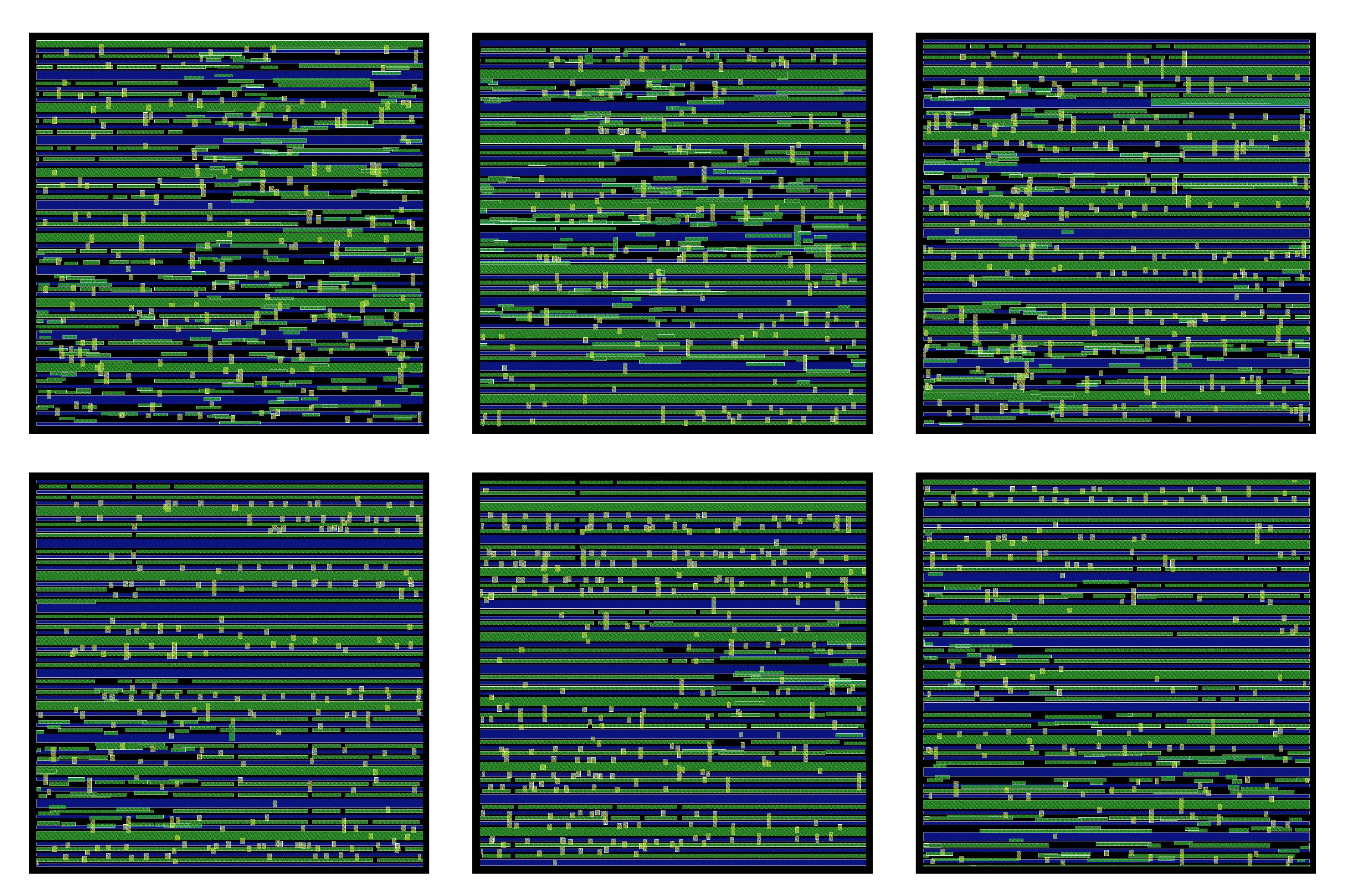}
    \caption{\textbf{Qualitative results on ContLayNet.} IPAM produces more balanced and well-distributed layers than baseline methods, offering opportunities for future improvement.}
    \label{fig:appendix_gds}
    \end{center}
\vspace{-10pt}
\end{figure*}

\begin{figure*}[t!]
{\color{blue}
    \begin{center}
    \includegraphics[width=\linewidth]{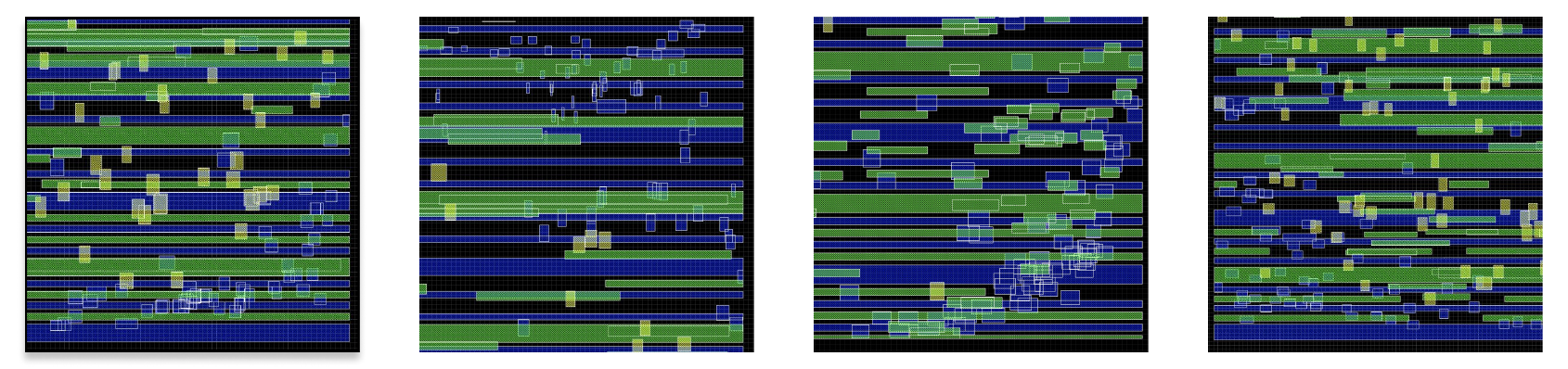}
    \caption{\textbf{Magnified DLT results on ContLayNet.} DLT generates numerous small, noisy elements due to poor length control in its non-autoregressive architecture.}
    \label{fig:appendix_magnified_DLT}
    \end{center}
}
\vspace{-10pt}
\end{figure*}

\cref{fig:appendix_gds} shows additional samples from IPAM, producing more balanced and well-distributed layers than baseline methods.
\cref{fig:appendix_magnified_DLT} shows magnified views of DLT outputs, revealing numerous small, noisy elements scattered throughout the layouts. This over-generation stems from DLT's lack of capability to control sequence termination in its non-autoregressive setting, which is particularly problematic for variable-length datasets like ContLayNet.
To further analyze the impact of our key components on length control, \cref{fig:ablation_gaussian} visualizes the distribution of length errors (sampled length $-$ ground truth length) by fitting Gaussian distributions to the errors. Our complete IPAM framework achieves near-zero mean error($\mu=0.15$) with reduced variance ($\sigma=146.13$), demonstrating accurate and consistent length prediction. In contrast, the ablated version without $\text{MLP}_{\text{EOS}}$ and $\mathcal{L}_\ell$ exhibits systematic bias ($\mu=38.91$) and higher variance ($\sigma=154.45$), indicating both accuracy and consistency issues. These results complement the quantitative ablation studies in \cref{tab:high_prec}, confirming that both EOS logit adjustment and length regularization are essential for precise length control in autoregressive generation.

\subsection{Practical Applications and Industrial Impact.}
\label{subsec:supp_impact}

The generated circuits from our approach offer several practical applications in semiconductor layout design.
Generated layouts provide manufacturable starting configurations for design engineers, substantially reducing design space exploration overhead while ensuring adherence to manufacturing constraints.
Our approach systematically identifies novel layout configurations that satisfy manufacturing requirements while exploring previously inaccessible regions of the feasible design space.
The model generates both compliant and non-compliant samples, enabling comprehensive analysis of constraint violations and potential failure modes critical for robust layout synthesis.

\section{Graphic Layout Generation}

\subsection{Empirical Analysis of Precision Limitations}
\cref{tab:app_analysis} demonstrates how precision affects LayoutTransformer~(LT) performance across both PubLayNet and Rico datasets.
The results confirm our theoretical analysis in~\cref{subsec:precision}: as precision increases, the exponential growth in token count leads to significant performance degradation.
At low precision~(4 bits), models exhibit poor performance with high FID scores due to insufficient representational capacity.
Performance peaks at moderate precision levels~(8-12 bits), where the model achieves an optimal balance between representational power and computational feasibility.
However, at higher precision~(16 bits), performance deteriorates markedly due to training instability caused by vocabulary explosion.
This pattern consistently appears across both datasets and all generation tasks, highlighting a fundamental limitation of discretization-based methods in high-precision settings.

\begin{table}[h]
\centering
\setlength{\tabcolsep}{3.5pt}
\caption{\textbf{Empirical Analysis of Precision Limitations.} Performance significantly degrades at higher levels~(16 bits) due to vocabulary explosion. Metrics: \textit{FID}~(lower is better), \textit{Overlap} and \textit{Alignment} scores~(closer to real is better, $\times 100$ for clarity).}
\label{tab:app_analysis}
\vspace{5pt}
\scalebox{0.85}{
\begin{tabular}{lcccccc|cccccc}
\toprule
Dataset & \multicolumn{6}{c|}{\textbf{PubLayNet}} & \multicolumn{6}{c}{\textbf{Rico}} \\ \midrule
Task & \multicolumn{3}{c|}{Completion} & \multicolumn{3}{c|}{Un-Gen} & \multicolumn{3}{c|}{Completion} & \multicolumn{3}{c}{Un-Gen} \\ \midrule
Precision & FID & Overlap & \multicolumn{1}{c|}{Align.} & FID & Overlap & Align. & FID & Overlap & \multicolumn{1}{c|}{Align.} & FID & Overlap & Align. \\ \midrule
4 & \ \ \ 49.25 & \ \ 5.25 & \multicolumn{1}{c|}{0.15} & \ \  55.49 & \ \ 4.91 & 0.17 & 87.13 & 33.78 & \multicolumn{1}{c|}{0.32} & 86.96 & 34.50 & 0.31 \\
8 & \ \ \ \ 3.00 & \ \ 2.78 & \multicolumn{1}{c|}{0.08} & \ \ 11.12 & \ \ 2.50 & 0.13 & \ \ 5.20 & 53.47 & \multicolumn{1}{c|}{0.21} & \ \ 6.00 & 52.40 & 0.20 \\
12 & \ \ 12.93 & \ \ 7.72 & \multicolumn{1}{c|}{0.13} & \ \ 22.51 & \ \ 8.69 & 0.19 & \ \ 8.90 & 58.22 & \multicolumn{1}{c|}{0.29} & 10.53 & 58.79 & 0.26 \\
16 & 165.63 & 57.08 & \multicolumn{1}{c|}{0.75} & 178.68 & 52.69 & 1.42 & 20.35 & 50.17 & \multicolumn{1}{c|}{0.39} & 22.08 & 51.44 & 0.30 \\ \midrule
Real & - & 0.22 & \multicolumn{1}{c|}{0.03} & - & 0.22 & 0.03 & - & 58.96 & \multicolumn{1}{c|}{0.17} & - & 58.96 & 0.17 \\ \bottomrule
\end{tabular}
}
\end{table}

\subsection{Experiment Details}
\label{subsec:supp_B2}
\paragraph{Datasets.}
\textit{PubLayNet}~\citep{zhong2019publaynetlargestdatasetdocument} contains 330K scientific document layouts with five component types: text, title, figure, list, and table.
\textit{Rico}~\citep{Deka:2017:Rico} comprises 91K mobile UI layouts with 27 component types.
We focused on the top 13 component types, following the approach of~\citep{lee2020neuraldesignnetworkgraphic}.
For both datasets, we limited our analysis to layouts containing 9 or fewer components, consistent with methods~\citep{Kikuchi2021constrainedgraphiclayoutgeneration, levi2023dltconditionedlayoutgeneration}.
We used the train-test split defined by~\citep{Kikuchi2021constrainedgraphiclayoutgeneration}.

\paragraph{Metrics.}
We evaluate layout quality using three common metrics: {Fr\'echet inception distance}~(FID)~\citep{heusel2018ganstrainedtimescaleupdate}, which measures the similarity between generated and real layouts; \textit{Overlap}~\citep{li2019layoutgangeneratinggraphiclayouts}, which quantifies the total overlapping area between components; and \textit{Alignment} score~\citep{lee2020neuraldesignnetworkgraphic}, which evaluates how well components are aligned.
We implement all of these metrics following LayoutGAN++~\citep{Kikuchi2021constrainedgraphiclayoutgeneration}, utilizing the same pre-trained feature extraction model.

\paragraph{Implementation details.}
We adopt a Transformer decoder architecture with a hidden dimension of 1024, 6 decoder layers, and 8 attention heads.
The denoising MLP consists of 6 blocks with 1024-channel width, and the diffusion process follows~\citep{nichol2021improveddenoisingdiffusionprobabilistic}.
During training, we sample the timestep $t$ 32 times for each latent vector $\mathbf{z}$ of the denoising MLP.
For PubLayNet's discrete branch and for EOS logit adjustment, we employ a two-layer MLP with GELU activation and dropout, using a hidden layer twice the input width.
Due to the training instability with smaller datasets, we use a simpler linear layer for Rico instead of the MLP.
The models were trained on PubLayNet~(learning rate $7.5 \times 10^{-5}$ and Rico~(learning rate $1.5 \times 10^{-4}$), both completing within 15 hours on a single RTX 4090.
We set $\lambda_1=100$, $\lambda_2=0.005$, and $\alpha=0.05$.

\subsection{Additional Results}

\begin{figure*}[t!]
    \begin{center}
    \includegraphics[width=\linewidth]{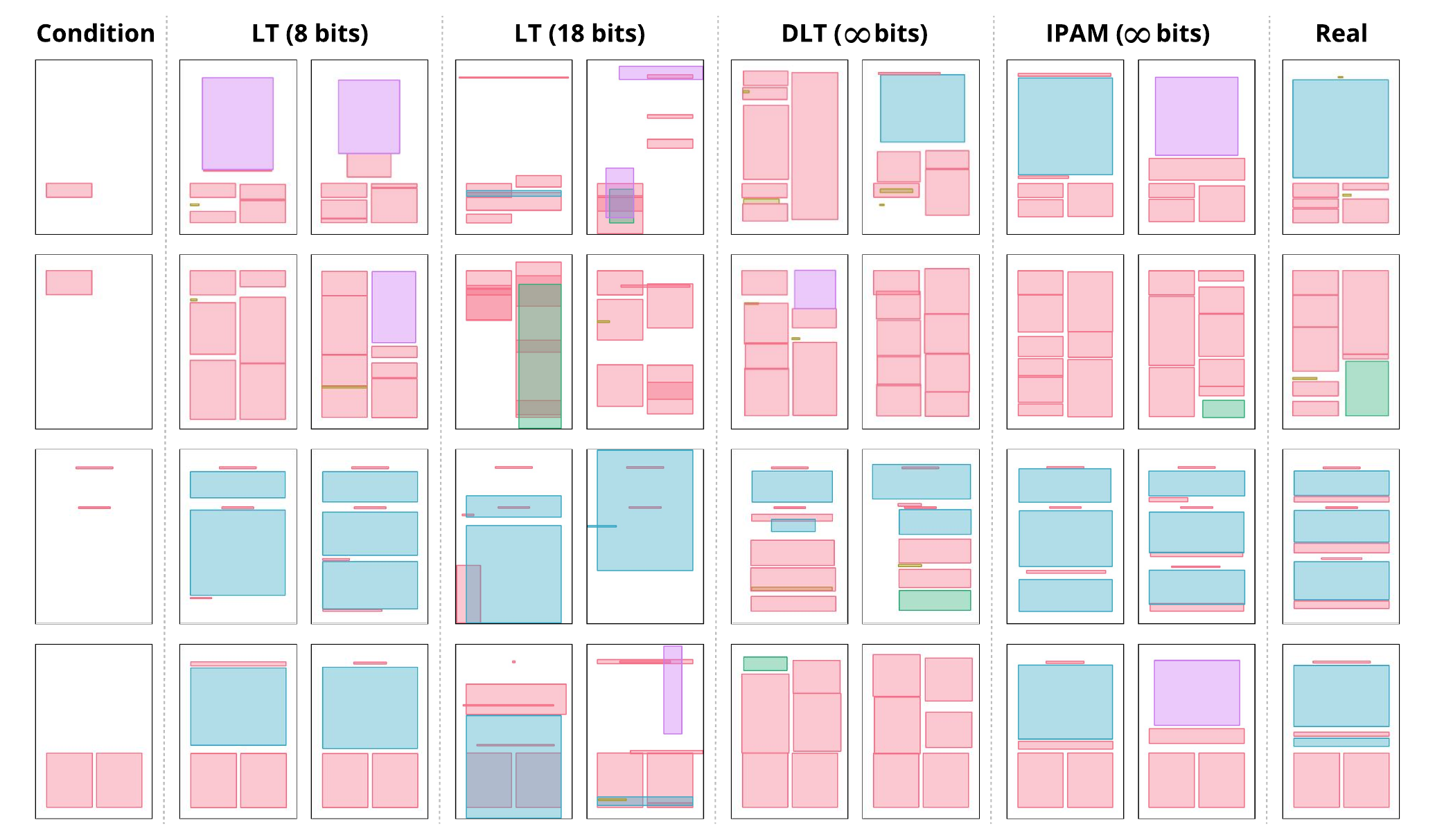}
    \caption{\textbf{Qualitative results on PubLayNet, Completion task.} IPAM~(Ours) achieves comparable quality to low-precision LT while outperforming high-precision LT and DLT.}
    \label{fig:lg_qual}
    \end{center}
\end{figure*}

\begin{figure*}[t!]
    \begin{center}
    \includegraphics[width=\linewidth]{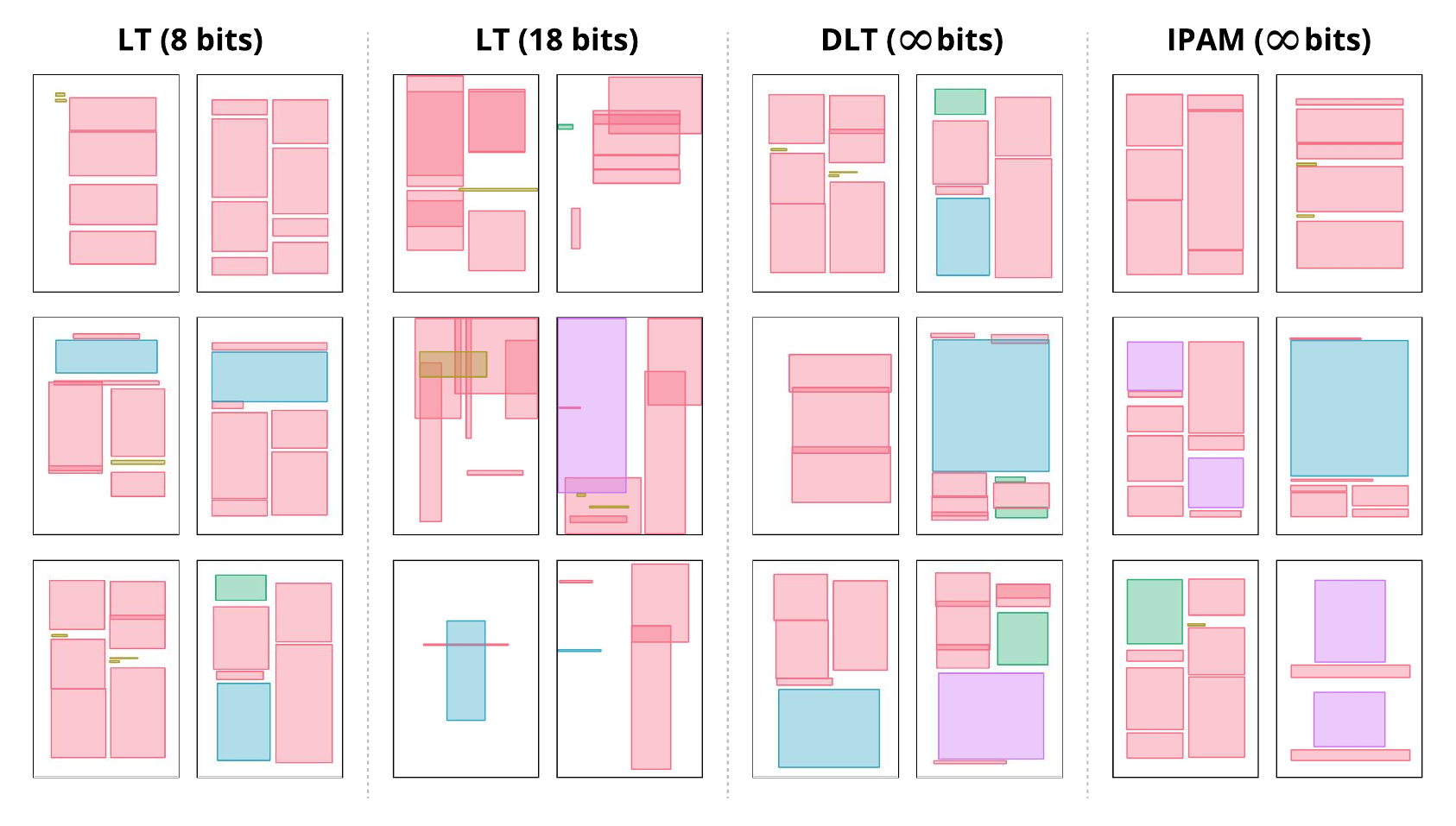}
    \caption{\textbf{Qualitative results on PubLayNet, Un-Gen task.} IPAM (Ours) achieves comparable quality to low-precision LT while outperforming high-precision LT and DLT.}
    \label{fig:appendix_publaynet_uncond}
    \end{center}
\end{figure*}

\paragraph{PubLayNet.}
\cref{fig:lg_qual} and \cref{fig:appendix_publaynet_uncond} present additional qualitative results for the completion task and unconditioned generation task on PubLayNet, respectively.
IPAM achieves comparable structural coherence to low-precision LT while maintaining higher precision capabilities.
Each rectangular box represents a region of a specific document element~(\textit{e.g.,} text, title, figure), with different colors indicating different element classes.

\begin{figure*}[t!]
    \begin{center}
    \includegraphics[width=\linewidth]{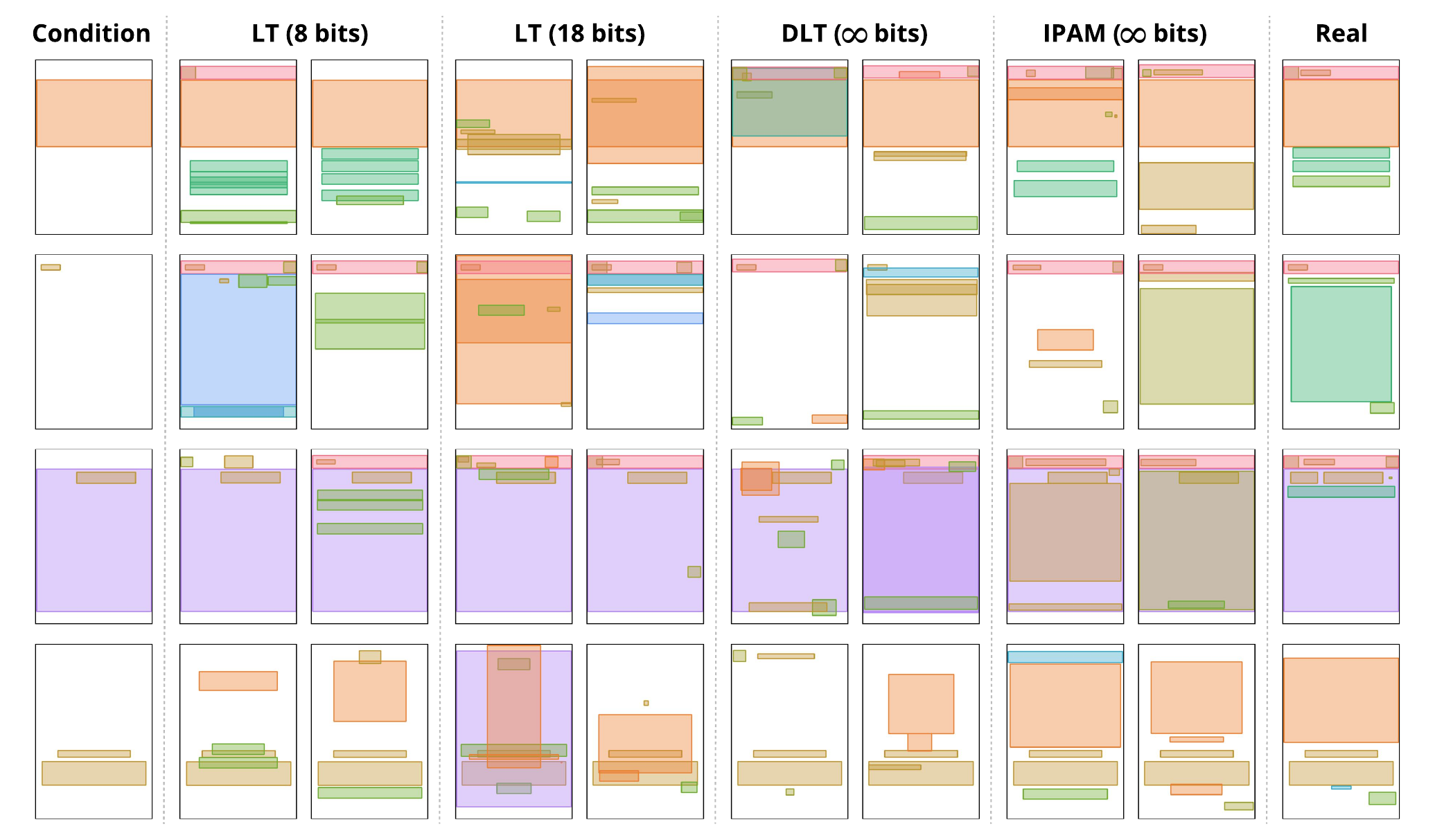}
    \caption{\textbf{Qualitative results on Rico, Completion task.} IPAM (Ours) achieves comparable quality to low-precision LT while outperforming high-precision LT and DLT.}
    \label{fig:appendix_rico_comp}
    \end{center}
\end{figure*}

\begin{figure*}[t!]
    \begin{center}
    \includegraphics[width=\linewidth]{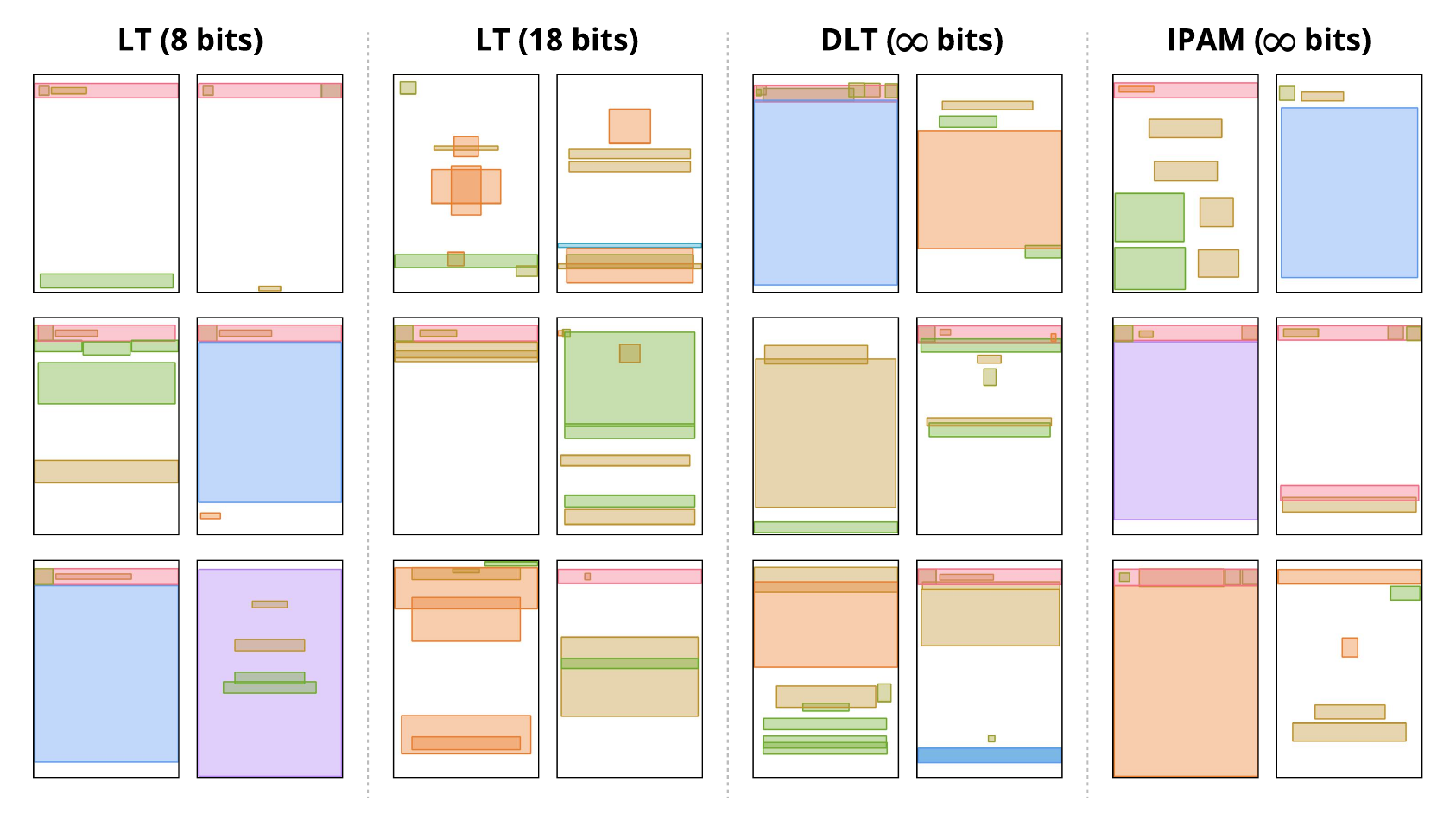}
    \caption{\textbf{Qualitative results on Rico, Un-Gen task.} IPAM (Ours) achieves comparable quality to low-precision LT while outperforming high-precision LT and DLT.}
    \label{fig:appendix_rico_uncond}
    \end{center}
\end{figure*}

\paragraph{Rico.}
\cref{fig:appendix_rico_comp} shows results for the completion task on Rico, where IPAM effectively preserves layout consistency and element relationships.
\cref{fig:appendix_rico_uncond} demonstrates unconditioned generation results, with generated layouts maintaining appropriate mobile UI design patterns.

\section{Text-to-SVG Generation}
\subsection{Experiment Details}
\label{subsec:supp_C1}

\paragraph{Text-to-SVG implementation.}
Following IconShop~\citep{wu2023iconshoptextguidedvectoricon}, text descriptions are tokenized using a pretrained BERT encoder and prepended to the SVG sequence for joint autoregressive modeling.
The model learns to map textual semantics to appropriate SVG commands and coordinates through end-to-end training, generating vector graphics that align with input text descriptions.

\paragraph{Datasets.}
\textit{FIGR-8-SVG}~\citep{clouatre2019figr} is composed of 1.5M samples of SVG-formatted black-and-white icons. The preprocessing of the data followed the procedure of Iconshop~\citep{wu2023iconshoptextguidedvectoricon} to get the valid text-SVG pairs. We excluded paired data with text longer than 50 words or SVG longer than 256 atomic units. For the training validation and testing of the models, 72K samples each are extracted from the data set as validation set and test set.

\paragraph{Metrics.} 
We evaluate SVG generation quality using two commonly adopted metrics: {Fr\'echet inception distance}~(FID)~\citep{heusel2018ganstrainedtimescaleupdate}, which measures the similarity between the generated SVG and real SVG in rendered image space and \textit{CLIP score}~\citep{radford2021learning}, which evaluates the similarity score between the generated rendered SVG icon and input text condition.

\paragraph{SVG representation.}

Following IconShop~\citep{wu2023iconshoptextguidedvectoricon}, each SVG atomic unit is represented with a standardized 8-dimensional continuous vector~(four coordinate pairs: start point, control point 1, control point 2, and end point).
Different command types use only relevant coordinates and ignore unused ones.
This unified representation enables consistent tensor operations across all command types while handling their varying coordinate requirements.

\paragraph{Implementation details.}
We adopt a Transformer decoder architecture with 8 decoder layers, 8 attention heads, and a hidden dimension of 1024.
For the denoising process, we employ an MLP comprising three blocks, each with 256 channels, following the diffusion approach of~\citep{nichol2021improveddenoisingdiffusionprobabilistic}.
During training, each latent vector $\mathbf{z}$ is processed through the denoising MLP, sampling the timestep $t$ 4 times each.
The discrete branch for SVG generation consists of a two-layer MLP with a hidden layer same as the input width, using RELU activation and dropout.
We set $\lambda_1 = 100$, $\lambda_2 = 0.001$, and $\alpha = 0.005$.
The model was trained on FIGR-8-SVG for 10 days on 4 $\times$ NVIDIA A6000 GPUs, with a learning rate of $1.0 \times 10^{-4}$.

\clearpage

\clearpage

\end{document}